\pdfoutput=1

\documentclass[11pt]{article}

\usepackage{acl}

\usepackage{times}
\usepackage{latexsym}

\usepackage[T1]{fontenc}

\usepackage[utf8]{inputenc}

\usepackage{microtype}

\usepackage{url}
\usepackage{CJKutf8}
\usepackage{fancyhdr}
\usepackage{amsmath}
\usepackage{amsfonts}
\usepackage{pgfplots}
\usepackage{subfigure}
\usepackage{tikz}
\usepackage{booktabs}
\usepackage{xcolor}
\usepackage{MnSymbol}
\usepackage{paralist}
\usepackage{pifont}
\usepackage{multirow}
\usepackage{array}
\usepackage{marvosym}
\usepackage{soul}
\usepackage{contour}
\usepackage{colortbl}
\usepackage{tabularx}
\usepackage{ifsym}
\usepackage{footmisc}
\usepackage{xurl}

\pgfplotsset{compat=1.17}
\usepgfplotslibrary{statistics}
\usetikzlibrary{fit, calc, decorations.pathreplacing, positioning, shapes.geometric, backgrounds, arrows.meta, shadings, patterns, math, matrix}
\usepgfplotslibrary{fillbetween}

\definecolor{frameworkpeach}{RGB}{255,239,213}
\definecolor{frameworkgreen}{RGB}{228,242,220}
\definecolor{frameworkblue}{RGB}{86,152,195}

\definecolor{lightgreen}{RGB}{228,242,220}
\definecolor{darkgreen}{RGB}{99,174,75}
\definecolor{frameworkblue}{RGB}{58,125,168}
\definecolor{darkred}{RGB}{191,38,50}

\definecolor{gray1}{RGB}{90,90,90}
\definecolor{gray2}{RGB}{150,150,150}

\newcommand\revised[1]{\textcolor{black}{#1}}

\providecommand{\e}[1]{\ensuremath{\times 10^{#1}}}

\newcommand{\minitab}[2]{\renewcommand{\arraystretch}{0.5}\begin{tabular}{m{#1}<{\centering}}#2\end{tabular}}
\newcommand\resultbox[2]{\minitab{34pt}{#1\\\sd{#2}}}
\newcommand\resultboxud[2]{\minitab{30pt}{#1\\\sd{#2}}}
\newcommand\namebox[1]{#1}
\newcommand\bt[1]{\textbf{\ul{#1}}}
\newcommand\lf[1]{\textcolor{gray1}{#1}}
\newcommand\ls[1]{\textcolor{gray2}{#1}}
\newcommand\s{\ls{s}}
\setuldepth{A}
\renewcommand\subparagraph[1]{\noindent $\bullet$ \ul{#1}}
\newcommand\topsh{\rule{0pt}{0.8\normalbaselineskip} }

\title{Bridging Pre-trained Language Models and Hand-crafted Features \\for Unsupervised {POS} Tagging}

\author{
Houquan Zhou, Yang Li\Thanks{$~$ Houquan and Yang make equal contributions to this
work. Zhenghua is the corresponding author.}, Zhenghua Li$^{\text{\Letter}}$\kern-0.6em,\, Min Zhang \\
Institute of Artificial Intelligence, School of Computer Science and Technology, \\Soochow University, China \\
{\tt \{hqzhou,ylinlp\}@stu.suda.edu.cn}; 
{\tt \{zhli13,minzhang\}@suda.edu.cn} \\
}

\begin{document}
\begin{CJK*}{UTF8}{gbsn}
\maketitle
\begin{abstract}
        In recent years, large-scale pre-trained language models (PLMs) have made extraordinary progress in most NLP tasks. 
        But, in the unsupervised POS tagging task, works utilizing PLMs are few and fail to achieve state-of-the-art (SOTA) performance.
        The recent SOTA performance is yielded by a Guassian HMM variant proposed by \citet{he2018unsupervised}.
        However, as a generative model, HMM makes very strong independence assumptions, making it very challenging to incorporate contexualized word representations from PLMs. 
        In this work, we for the first time propose a neural conditional random field autoencoder (CRF-AE) model for unsupervised POS tagging. 
        The discriminative encoder of CRF-AE can straightforwardly incorporate PLM word representations.
        Moreover, inspired by feature-rich HMM, we reintroduce hand-crafted features into the decoder of CRF-AE. 
        Finally, experiments clearly show that our model outperforms previous state-of-the-art models by a large margin on Penn Treebank and multilingual Universal Dependencies treebank v2.0. 
\end{abstract}

\section{Introduction}

\begin{figure}[tb]
    \centering
    \begin{tikzpicture}[
        anchor point/.style={
            draw=none,
            minimum height=0.4cm
        },
        word/.style={
            anchor=base,
        },
    ]
        \foreach \x in {0,...,5}
        {
            \node[anchor point] (x_\x) at (${\x-2.5}*(1.25cm, 0)$) {\phantom{AAA}};
            \node[anchor point] (y_\x) at (${\x-2.5}*(1.25cm, 0) + (0, 0.8)$) {\phantom{\texttt{AAA}}};
        }
        
        \node[word] (w_0) at (x_0.base) {I};
        \node[word] (w_1) at (x_1.base) {looked};
        \node[word] (w_2) at (x_2.base) {at};
        \node[word] (w_3) at (x_3.base) {my};
        \node[word] (w_4) at (x_4.base) {watch};
        \node[word] (w_5) at (x_5.base) {.};
        
        \node[word] (t_0) at (y_0.base) {\texttt{PRP}};
        \node[word] (t_1) at (y_1.base) {\texttt{VBD}};
        \node[word] (t_2) at (y_2.base) {\texttt{IN}};
        \node[word] (t_3) at (y_3.base) {\texttt{PRP\$}};
        \node[word] (t_4) at (y_4.base) {\texttt{NN}};
        \node[word] (t_5) at (y_5.base) {\texttt{.}};
        
        \foreach \x in {0,...,5}
        {
            \draw[thin, gray] (x_\x) -- (y_\x);
        }
        
        \begin{scope}[on background layer]
            \node [fill=lightgray!10, draw=black, rounded corners=2mm, thick, fit=(x_0) (y_0) (x_5) (y_5)] (bg) {};
        \end{scope}
        
    \end{tikzpicture}
    \caption{Example of POS tagging.}
    \label{fig:example}
\end{figure}
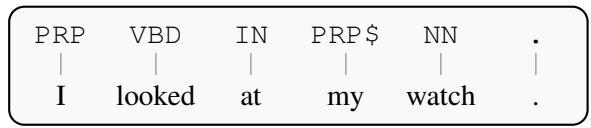

Unsupervised learning has been an important yet challenging research direction in NLP \cite{klein-etal-2004-corpus,liang-etal-2006-alignment,seginer-2007-fast}. Training models directly from unlabeled data can relieve painful data annotation and is thus especially attractive for low-resource languages \cite{he2018unsupervised}. 
As three typical tasks related to syntactic analysis, unsupervised part-of-speech (POS) tagging (or induction), dependency parsing, and constituency parsing have attracted intensive interest during the past three decades \cite[\textit{inter alia}]{pereira-schabes-1992-inside-outside,christodoulopoulos-etal-2010-two}. 
Compared with tree-structure dependency and constituency parsing, POS tagging corresponds to simpler sequential structure, and aims to assign a POS tag to each word, as depicted in Figure \ref{fig:example}. 
Besides the alleviation of labeled data, unsupervised POS tagging is particularly valuable for child language acquisition study because every child manages to induce syntactic categories without access to labeled data \cite{yuret-etal-2014-unsupervised}.

Nowadays, supervised POS tagging models trained on large-scale labeled data can already achieve extremely high accuracy, for example over 97.5\% on English Penn Treebank (PTB) texts \citep{huang-etal-2015-bidirectional, bohnet-etal-2018-morphosyntactic, zhou-etal-2020-is}.  
However, unsupervised POS tagging, though having attracted a lot of research interest \citep{lin2015unsupervised, tran2016unsupervised, he2018unsupervised, stratos2019mutual, gupta2020clustering}, can only achieve at most 80.8\% many-to-one (M-1) accuracy, where M-1 means multiple induced tags can be mapped to a single ground-truth tag when evaluating the model on the test data. 

The generative Hidden Markov Models (HMMs) are the most representative and successful approach for unsupervised POS tagging \citep{merialdo-1994-tagging, ganchev-etal-2009-posterior}. 
By treating POS tags as latent variables, a first-order HMM factorizes the joint probability of a sentence and a tag sequence $p(\mathbf{x}, \mathbf{y})$ into independent emission probabilities $p(x_i\mid y_i)$ and transition probabilities $p(y_{i-1}\mid y_i)$. The training objective is to maximize the marginal probability $p(\mathbf{x})$, which can be solved by the EM algorithm or direct gradient descent \citep{salakhutdinov-etal-2003-optimization}. 
\citet{berg2010painless} propose a feature-rich HMM (FHMM), which further parameterizes $p(x_i\mid y_i)$ with many hand-crafted morphological features, greatly boosting M-1 accuracy to 75.5 from 63.1 of the basic HMM.

In the DL era, researchers have paid a lot of attention to HMMs for unsupervised POS tagging. 
\citet{lin2015unsupervised} propose a Gaussian HMM (GHMM), where $p(x_i\mid y_i)$ corresponds to the probability of the pre-trained word embedding (fixed during training) of $x_i$ against the Gaussian distribution of $y_i$.
\citet{tran2016unsupervised} propose a neural HMM model (NHMM), where  $p(x_i\mid y_i)$ and $p(y_{i-1}\mid y_i)$ are all computed via neural networks with POS tag and word embeddings as inputs. 
\citet{he2018unsupervised} extend the Gaussian HMM of \citet{lin2015unsupervised} by introducing an invertible neural projection (INP) component for the pre-trained word embeddings, which has a similar effect of tuning word embeddings during training. 
Their INP Gaussian HMM (INP-GHMM) approach %
achieves state-of-the-art (SOTA) M-1 accuracy (80.8) on PTB so far. 

The major weakness of HMMs is the strong independence assumption in emission probabilities $p(x_i\mid y_i)$, which directly hinders the use of contextualized word representations from powerful pre-trained language models (PLMs) such as ELMo/BERT \citep{peters-etal-2018-deep, devlin-etal-2019-bert}. 
It is a pity since PLMs are able to greatly boost performance of many NLP tasks. 

In this work, we for the first time propose a neural conditional random field autoencoder (CRF-AE) model for unsupervised POS tagging, inspired by \citet{ammar2014conditional} who propose a non-neural CRF-AE model. 
In the discriminative encoder of CRF-AE, we straightforwardly incorporate ELMo word representations. 
Moreover, inspired by feature-rich HMM \citep{berg2010painless}, we reintroduce hand-crafted features into the decoder of CRF-AE. 
In summary, this work makes the following contributions:
\begin{compactitem}[$\bullet$]
    \item We for the first time propose a neural CRF-AE model for unsupervised POS tagging. 
    \item We successfully bridge PLMs and hand-crafted features in our CRF-AE model. 
    \item Our model achieves new  SOTA M-1 accuracy of 83.21 on the 45-tag English PTB data and outperforms the previous best result by 2.41. 
    \item After a few straightforward adjustments, our model achieves new SOTA M-1 accuracy on the 12-tag  multilingual Universal Dependencies treebank v2.0 (UD), surpassing the previous best results by 4.97 on average. 
\end{compactitem}
We release our code at \url{https://github.com/Jacob-Zhou/FeatureCRFAE}, including our re-implemented HMM and FHMM models.

\section{Vanilla CRF-AE} %
\begin{figure}[tb]
    \centering
    \begin{tikzpicture}[
        connect/.style={
                semithick,
                shorten >= 1.5pt,
                shorten <= 1.5pt,
            },
        arrow/.style={
                arrows = {-Straight Barb[length=0.8mm]},
                shorten >= 2pt,
                shorten <= 1.5pt,
                semithick,
                rounded corners=2pt,
            },
        small_arrow/.style={
                arrows = {-Straight Barb[length=0.4mm]},
                shorten >= 0pt,
                shorten <= 1.5pt,
                draw=gray,
                thin
            },
        input/.style={
                draw=none,
                minimum height=0.4cm,
            },
        module/.style={
                solid,
                draw,
                minimum height=0.5cm,
                rectangle,
                rounded corners=2mm,
                thick,
                align=center,
                fill=none
            },
        label/.style={
                anchor=east,
                draw=none,
                fill=white,
                rectangle,
                inner sep=4pt,
            },
        prob_node/.style={
                draw=black,
                fill=lightgray!15,
                solid,
                circle,
                thick,
                anchor=south,
                minimum size=0.6cm
            },
        prob_para/.style={
                solid,
                draw,
                fill=black,
                minimum size=0.12cm,
                inner sep=0pt,
            },
        repr/.style={
                solid,
                draw,
                thin,
                anchor=center, 
                minimum size=0.18cm,
                inner sep=0pt,
            },
        repr_placeholder/.style={
                minimum size=0.18cm, 
                anchor=south, 
                inner sep=0pt
            }
        ]
        \centering

        \node[prob_node] (bx) at (0, 0) {};
        \node[anchor=base] at ($(bx.base) + (0.005, -0.08)$) {\footnotesize $\mathbf{x}$};

        \foreach \x in {0,...,4}
        {
            
            \node[prob_node] (y_\x) at (${\x-2}*(1.4cm, 0) + (0, 1.25cm)$) {};
            \draw[connect] (bx) -- (y_\x);
            
            \node[prob_para] at ($(bx)!0.5!(y_\x)$) {};
            \node[prob_node] (x_\x) at ($(y_\x.north) + (0, 0.45cm)$) {};
            \draw[arrow] (y_\x) -- (x_\x);
        }
       
        \foreach \x in {0,...,3}
        {
            \tikzmath{
                int \xp;
                \xp = \x+1;
            }
            \node[prob_para] (pp_\x) at ($(y_\x.north)!0.5!(y_\xp.south)$) {};
            \draw[connect] (y_\x) -- (y_\xp);
        }

        \node[anchor=base] at ($(y_0.base) + (0.02, -0.05)$) {\footnotesize $y_1$};
        \node[anchor=base] at ($(y_1.base) + (0.02, -0.05)$) {\footnotesize $y_2$};
        \node[anchor=base] at ($(y_2.base) + (0.02, -0.05)$) {\footnotesize $y_3$};
        \node[anchor=base] at ($(y_3.base) + (0, -0.05)$) {\footnotesize $\ldots$};
        \node[anchor=base] at ($(y_4.base) + (0.02, -0.05)$) {\footnotesize $y_n$};
        
        \node[anchor=base] at ($(x_0.base) + (0.02, -0.06)$) {\footnotesize $x_1$};
        \node[anchor=base] at ($(x_1.base) + (0.02, -0.06)$) {\footnotesize $x_2$};
        \node[anchor=base] at ($(x_2.base) + (0.02, -0.06)$) {\footnotesize $x_3$};
        \node[anchor=base] at ($(x_3.base) + (0, -0.05)$) {\footnotesize $\ldots$};
        \node[anchor=base] at ($(x_4.base) + (0.02, -0.06)$) {\footnotesize $x_n$};

    \end{tikzpicture}
    \caption{
      Illustration of CRF-AE.
    }
    \label{fig:crfae}
  \end{figure}
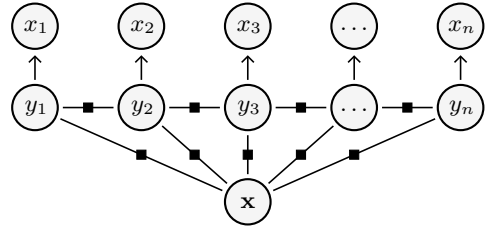
\label{sec:vanilla-crf-ae}
In this work, we adopt the CRF-AE approach as our basic model for unsupervised POS tagging. 
The non-neural CRF-AE model is first proposed by 
\citet{ammar2014conditional} for unsupervised sequence labeling tasks, inspired by neural network autoencoders. \citet{cai-etal-2017-crf} also extend the idea to  non-neural unsupervised dependency parsing. 
The basic idea is first producing latent structures, i.e., POS tag sequences, with a discriminative CRF over the observed sentence, and then reconstructing the original sentence given each latent structure. 
The two steps correspond to the encoder and the decoder respectively. 

\textbf{Training loss.} We denote a sentence as $\mathbf{x}=x_1, x_2, \cdots, x_i, \cdots, x_n$, and a POS tag sequence as $\mathbf{y}=y_1, y_2, \cdots, y_i, \cdots, y_n$. Given an unlabeled dataset $\mathcal{D}$ which does not contain any POS tag sequences, the training loss is:
    \begin{align}
        \mathcal{L}(\mathcal{D}; \boldsymbol{\phi},\boldsymbol{\theta}) =& - \sum_{\mathbf{x} \in \mathcal{D}}{\log{\mathbb{E}_{\mathbf{y}\sim p(\mathbf{y}\mid \mathbf{x}; \boldsymbol{\phi})}p(\mathbf{x}\mid \mathbf{y}; \boldsymbol{\theta})}} \nonumber\\
        &+ \lambda\left(\left\| \boldsymbol{\phi}\right\| ^2_2 + \left\| \boldsymbol{\theta}\right\| ^2_2\right),
    \end{align}
where $p(\mathbf{y}\mid \mathbf{x}; \boldsymbol{\phi})$ is the CRF encoder; $p(\mathbf{x}\mid \mathbf{y}; \boldsymbol{\theta})$ is the decoder;  $\boldsymbol{\phi}$ and $\boldsymbol{\theta}$ are model parameters. 

This training loss encourages the model to meet the intuition that 
a high-probability POS sequence should also permit reconstruction of the sentence with a high probability.

\citet{ammar2014conditional} adopt the \textit{Expectation-Maximization} (EM) algorithm for training. In this work, we directly compute the training loss via the \textit{Forward} algorithm.
Then, we employ the powerful \textit{AutoGrad} function of deep learning to compute the gradient of each parameter. Our preliminary experiments on HMM and feature-rich HMM show that this gradient-based approach is consistently superior to EM in both efficiency and performance.

\textbf{Inference}. During evaluation, we follow \citet{ammar2014conditional} and use both the CRF and the reconstruction probabilities to obtain the optimal tag sequence:
\begin{equation}
    \mathbf{y}^\ast = \arg \max_{\mathbf{y}}
    p(\mathbf{y}\mid \mathbf{x}; \boldsymbol{\phi}) p(\mathbf{x}\mid \mathbf{y}; \boldsymbol{\theta}),
\end{equation}
which can be solved by the \textit{Viterbi} algorithm.

\textbf{CRF Encoder}: $p(\mathbf{y}\mid \mathbf{x}; \boldsymbol{\phi})$. 
As a discriminative log-linear model, the CRF encoder defines a conditional probability:
\begin{equation}
    \begin{split}
        p(\mathbf{y}\mid \mathbf{x}; \boldsymbol{\phi}) &= \frac{\exp{\left(S(\mathbf{x}, \mathbf{y}; \boldsymbol{\phi})\right)}}{Z(\mathbf{x}; \boldsymbol{\phi})\equiv\sum_{\mathbf{y}}\exp({S(\mathbf{x}, \mathbf{y}; \boldsymbol{\phi}))}}, \\
    \end{split}
\end{equation}
where $Z(\mathbf{x})$ is the partition function, also known as the normalization term.

The score of $\mathbf{y}$ given $\mathbf{x}$ is decomposed into bigram scores: 
\begin{equation} \label{equ:sequence-score}
    S(\mathbf{x}, \mathbf{y}; \boldsymbol{\phi}) = \sum_{i=1}^{n}{s\left(\mathbf{x}, y_{i-1}, y_i; \boldsymbol{\phi}\right)}.
\end{equation}

\citet{ammar2014conditional} use hand-crafted discrete features to obtain bigram scores.
\begin{equation}
    s\left(\mathbf{x}, y_{i-1}, y_i; \boldsymbol{\phi}\right) = \boldsymbol{\phi}^\top g(\mathbf{x}, y_{i-1}, y_i, i).
\end{equation}

\textbf{Decoder}: $p(\mathbf{x}\mid \mathbf{y}; \boldsymbol{\theta})$. 
The decoder computes the reconstruction probability of $\mathbf{x}$ given a POS tag sequence $\mathbf{y}$, which %
is factorized into position-wise generation probabilities based on a strong independence assumption. 
\begin{equation} \label{equ:original-decoder}
    p(\mathbf{x}\mid \mathbf{y}; \boldsymbol{\theta}) = \prod_{i=1}^{n}p(x_i\mid y_i; \boldsymbol{\theta}).
\end{equation}
\citet{ammar2014conditional} use a categorical distribution matrix $\boldsymbol{\theta}$, which is updated via EM training, to maintain all generation probabilities $p(x_i\mid y_i)$, i.e., a word $x_i$ generated by a tag $y_i$. 
\begin{figure}[tb]
    \centering
    \begin{tikzpicture}[
        connect/.style={
                semithick,
                shorten >= 1.5pt,
                shorten <= 1.5pt,
            },
        arrow/.style={
                arrows = {-Straight Barb[length=0.8mm]},
                shorten >= 2pt,
                shorten <= 1.5pt,
                semithick,
                rounded corners=2pt,
            },
        small_arrow/.style={
                arrows = {-Straight Barb[length=0.4mm]},
                shorten >= 0pt,
                shorten <= 1.5pt,
                draw=gray,
                thin
            },
        input/.style={
                draw=none,
                minimum height=0.4cm,
            },
        module/.style={
                solid,
                draw,
                minimum height=0.5cm,
                rectangle,
                rounded corners=2mm,
                thick,
                align=center,
                fill=none
            },
        label/.style={
                anchor=east,
                draw=none,
                fill=white,
                rectangle,
                inner sep=4pt,
            },
        prob_node/.style={
                draw=black,
                fill=lightgray!15,
                solid,
                circle,
                thick,
                anchor=south,
                minimum size=0.6cm
            },
        prob_para/.style={
                solid,
                draw,
                fill=black,
                minimum size=0.12cm,
                inner sep=0pt,
            },
        repr/.style={
                solid,
                draw,
                thin,
                anchor=center, 
                minimum size=0.18cm,
                inner sep=0pt,
            },
        repr_placeholder/.style={
                minimum size=0.18cm, 
                anchor=south, 
                inner sep=0pt
            }
        ]
        \centering
        \foreach \x in {0,...,4}
            \node[input] (input_\x) at (${\x-2}*(1.6cm, 0)$) {};
            
        \node[anchor=base] at (input_0.base) {\footnotesize \footnotesize $x_1$};
        \node[anchor=base] at (input_1.base) {\footnotesize \footnotesize $x_2$};
        \node[anchor=base] at (input_2.base) {\footnotesize \footnotesize $x_3$};
        \node[anchor=base] at (input_3.base) {\footnotesize \footnotesize $\ldots$};
        \node[anchor=base] (input_x_n) at (input_4.base) {\footnotesize \footnotesize $x_n$};
  
        \node [module, fill=frameworkpeach!50] [minimum width=7.4cm, minimum height=0.6cm, anchor=south] (plm) at (0, 0.65cm) {\footnotesize \texttt{Pre-trained Language Model}};

        \foreach \x in {0,...,4}
        {
            \foreach \y in {0,...,3}
            {
                \node[repr_placeholder] (repr_\x_\y) at (${\x-2}*(1.6cm, 0) + {\y}*(0, 0.16cm) + (plm.north) + (0, 0.35cm)$) {};
                \tikzmath{
                    if \y<2 then { let \c = $\filledtriangleleft$; let \p1 = west; let \p2 = east; } else { let \c = $\filledtriangleright$; let \p2 = west; let \p1 = east; };
                }
                \node[repr_placeholder, minimum size=0.08cm, anchor=\p2] at ($(repr_\x_\y.\p1)$) {\scriptsize \c};
            }
        }

        \node[repr, fill=darkred!6]  at (repr_0_3) {};
        \node[repr, fill=frameworkblue!80] at (repr_0_2) {};
        \node[repr, fill=darkred!65]  at (repr_0_1) {};
        \node[repr, fill=darkred!77]  at (repr_0_0) {};
        
        \node[repr, fill=darkred!95]  at (repr_1_3) {};
        \node[repr, fill=frameworkblue!85] at (repr_1_2) {};
        \node[repr, fill=darkred!70] at (repr_1_1) {};
        \node[repr, fill=darkred!75]  at (repr_1_0) {};
        
        \node[repr, fill=darkred!82] at (repr_2_3) {};
        \node[repr, fill=frameworkblue!75] at (repr_2_2) {};
        \node[repr, fill=frameworkblue!6] at (repr_2_1) {};
        \node[repr, fill=darkred!79]  at (repr_2_0) {};
        
        \node[repr, fill=darkred!60]  at (repr_3_3) {};
        \node[repr, fill=darkred!75]  at (repr_3_2) {};
        \node[repr, fill=darkred!95] at (repr_3_1) {};
        \node[repr, fill=darkred!85] at (repr_3_0) {};
        
        \node[repr, fill=darkred!66] at (repr_4_3) {};
        \node[repr, fill=darkred!2]  at (repr_4_2) {};
        \node[repr, fill=frameworkblue!4] at (repr_4_1) {};
        \node[repr, fill=darkred!75]  at (repr_4_0) {};

        \foreach \x in {0,...,4}
        {
            \foreach \y in {0,...,3}
            { 
                \node[repr_placeholder] (minus_\x_\y) at (${\y}*(0, 0.16cm) + (repr_\x _3.north) + (0, 0.3cm)$) {};
            }
        }
            
        \node[repr, fill=darkred!6]  at (minus_0_3) {};
        \node[repr, fill=frameworkblue!80] at (minus_0_2) {};
        \node[repr, fill=frameworkblue!5]   at (minus_0_1) {};
        \node[repr, fill=darkred!2]   at (minus_0_0) {};

        \node[repr, fill=darkred!89]   at (minus_1_3) {};
        \node[repr, fill=frameworkblue!5]  at (minus_1_2) {};
        \node[repr, fill=darkred!76] at (minus_1_1) {};
        \node[repr, fill=frameworkblue!4]  at (minus_1_0) {};

        \node[repr, fill=frameworkblue!13] at (minus_2_3) {};
        \node[repr, fill=darkred!10]  at (minus_2_2) {};
        \node[repr, fill=frameworkblue!101] at (minus_2_1) {};
        \node[repr, fill=frameworkblue!6]  at (minus_2_0) {};

        \node[repr, fill=frameworkblue!22]  at (minus_3_3) {};
        \node[repr, fill=darkred!150]  at (minus_3_2) {};
        \node[repr, fill=darkred!99] at (minus_3_1) {};
        \node[repr, fill=darkred!10] at (minus_3_0) {};

        \node[repr, fill=darkred!6] at (minus_4_3) {};
        \node[repr, fill=frameworkblue!73]  at (minus_4_2) {};
        \node[repr, fill=frameworkblue!4] at (minus_4_1) {};
        \node[repr, fill=darkred!75]  at (minus_4_0) {};

        \foreach \x in {0,...,4}
        {
            \foreach \y in {0,...,2}
                \node[repr_placeholder] (bn_\x_\y) at (${\y}*(0, 0.16cm) + (minus_\x _3.north) + (0, 0.65cm)$) {};
        }

        \node[repr, anchor=center, fill=darkred!7]  at (bn_0_1) {};
        \node[repr, anchor=center, fill=darkred!90]  at (bn_0_0) {};

        \node[repr, anchor=center, fill=darkred!85]  at (bn_1_1) {};
        \node[repr, anchor=center, fill=darkred!72]  at (bn_1_0) {};

        \node[repr, anchor=center, fill=frameworkblue!8]  at (bn_2_1) {};
        \node[repr, anchor=center, fill=frameworkblue!5]  at (bn_2_0) {};

        \node[repr, anchor=center, fill=darkred!80]  at (bn_3_1) {};
        \node[repr, anchor=center, fill=frameworkblue!15]  at (bn_3_0) {};

        \node[repr, anchor=center, fill=frameworkblue!2]  at (bn_4_1) {};
        \node[repr, anchor=center, fill=darkred!92]  at (bn_4_0) {};
        
        \foreach \x in {0,...,3}
        {
            \tikzmath{
                int \xp;
                \xp = \x+1;
            }
            
            \node[inner sep=0pt, circle] (mo_\x) at ($(repr_\x_3)!0.5!(minus_\xp_0)$) {$\ominus$};
            \draw[arrow, shorten >= 0pt, shorten <= 5.5pt] (repr_\x_2.south east) -- ($(mo_\x.south) - (0.25cm, 0.15cm)$) -- (mo_\x);
            \draw[arrow, shorten >= 0pt, shorten <= 5.5pt] (repr_\xp_2.south west) -- ($(mo_\x.south) - (-0.25cm, 0.15cm)$) -- (mo_\x);
            \draw[arrow, shorten <= 0pt, shorten >= 3.5pt] (mo_\x) -- ($(minus_\x_1) + (0.65cm, -0.1cm)$) -- (minus_\x_2);
            \draw[arrow, shorten <= 0pt, shorten >= 3.5pt] (mo_\x) -- ($(minus_\xp_1) + (-0.65cm, -0.1cm)$) -- (minus_\xp_2);
        }

        \foreach \x in {0,...,4}
        {
            \draw[arrow] (input_\x) -- (input_\x|-plm.south);
            \draw[arrow] (input_\x|-plm.north) -- (repr_\x_0);
            
            \draw[arrow] (minus_\x_3) -- (bn_\x_0);
        }

        \foreach \x in {0,...,4}
        {
            
            \node[prob_node] (y_\x) at ($(bn_\x_1.north) + (0, 0.5cm)$) {};
            \draw[connect] (bn_\x_1) -- (y_\x);
            
            \node[prob_para] at ($(bn_\x_1.north)!0.5!(y_\x.south)$) {};
            \node[prob_node] (x_\x) at ($(y_\x.north) + (0, 0.45cm)$) {};
            \draw[arrow] (y_\x) -- (x_\x);
            \node[inner sep=-0.6pt, fill=white] (mlp_\x) at ($(minus_\x_3.north)!0.5!(bn_\x_0.south) - (0, 0.04cm)$) {\huge $\vertbowtie$};
        }

        \foreach \x in {0,...,3}
        {
            \tikzmath{
                int \xp;
                \xp = \x+1;
            }
            \node[prob_para] (pp_\x) at ($(y_\x.north)!0.5!(y_\xp.south)$) {};
            \draw[connect] (y_\x) -- (y_\xp);
        }

        \node[anchor=base] at ($(y_0.base) + (0.02, -0.05)$) {\footnotesize $y_1$};
        \node[anchor=base] at ($(y_1.base) + (0.02, -0.05)$) {\footnotesize $y_2$};
        \node[anchor=base] at ($(y_2.base) + (0.02, -0.05)$) {\footnotesize $y_3$};
        \node[anchor=base] at ($(y_3.base) + (0, -0.05)$) {\footnotesize $\ldots$};
        \node[anchor=base] at ($(y_4.base) + (0.02, -0.05)$) {\footnotesize $y_n$};
        
        \node[anchor=base] at ($(x_0.base) + (0.02, -0.06)$) {\footnotesize $x_1$};
        \node[anchor=base] at ($(x_1.base) + (0.02, -0.06)$) {\footnotesize $x_2$};
        \node[anchor=base] at ($(x_2.base) + (0.02, -0.06)$) {\footnotesize $x_3$};
        \node[anchor=base] at ($(x_3.base) + (0, -0.05)$) {\footnotesize $\ldots$};
        \node[anchor=base] at ($(x_4.base) + (0.02, -0.06)$) {\footnotesize $x_n$};
        
        \node[inner sep=1pt, anchor=west] (p_label) at ($(x_0.south)!0.5!(y_0.north) + (0.1cm, 0.1cm)$) {\scriptsize $p(x_1|y_1; \boldsymbol{\theta})$};

        \node[inner sep=1pt, anchor=west] at ($(bn_0_1.north)!0.5!(y_0.south) + (0.1cm, 0.02cm)$) {\scriptsize $s\left(\mathbf{x}, y_1; \boldsymbol{\phi}\right)$};

        \node[inner sep=1pt, anchor=north] (t_label) at ($(y_1)!0.5!(y_2) - (0, 0.28)$) {\scriptsize $t\left(y_2, y_3; \boldsymbol{\phi}\right)$};
        
        \draw[small_arrow, -, shorten <= -2.0pt, shorten >= 1.5pt] (t_label) to[out=90, in=-90] (pp_1);

        \node[inner sep=1pt, anchor=east] at ($(repr_0_3.west)$) {\scriptsize $\overrightarrow{\mathbf{r}_{\!1}}$};
        \node[inner sep=1pt, anchor=west] at ($(repr_0_0.east)$) {\scriptsize $\overleftarrow{\mathbf{r}_{\!1}}$};
        
        \node[inner sep=1pt, anchor=base west, align=left] at ($(mo_0.base east) + (0, 0.02cm)$) {\scriptsize Minus Op.};

        \node[inner sep=1pt, anchor=west, align=left] at ($(mlp_0.east) - (0cm, -0.02cm)$) {\scriptsize Bottleneck \\[-5.5pt] \scriptsize MLP };
        
        \node[inner sep=1pt, anchor=west] at ($(minus_0_2.north east)$) {\scriptsize $\mathbf{m}_1$};
        
        \node[inner sep=1pt, anchor=west] at ($(bn_0_1.east) - (0, 0.008cm)$) {\scriptsize $\mathbf{c}_1$};

        \begin{scope}[on background layer]
            \node [fill=lightgray!2, draw=black, rounded corners=5mm, thin, densely dashed, fit=(y_0) (plm) (input_x_n)] (bg_encoder) {};
        \end{scope}

        \node[inner sep=1pt, anchor=south west] (cf_0) at ($(bg_encoder.north east|-x_4.north) + (-1.653cm, 0.3cm)$) {\scriptsize Capitalized: \ding{51}};
        \node[inner sep=1pt, anchor=south west] (cf_1) at ($(cf_0.north west) + (0, -0.05cm)$) {\scriptsize Capitalized: \ding{51}};
        \node[inner sep=1pt, anchor=south west] (cf_2) at ($(cf_1.north west) + (0, -0.05cm)$) {\scriptsize Capitalized: \ding{55}};
        \node[inner sep=1pt, anchor=south west] (cf_3) at ($(cf_2.north west) + (0, 0.18cm)$) {\scriptsize Capitalized: \ding{55}};

        \node[inner sep=1pt, anchor=base east] (df_0) at ($(cf_0.base west) + (-0.06cm, 0cm)$) {\scriptsize $\ldots$};
        \node[inner sep=1pt, anchor=base east] (df_1) at ($(cf_1.base west) + (-0.06cm, 0cm)$) {\scriptsize $\ldots$};
        \node[inner sep=1pt, anchor=base east] (df_2) at ($(cf_2.base west) + (-0.06cm, 0cm)$) {\scriptsize $\ldots$};
        \node[inner sep=1pt, anchor=base east] (df_3) at ($(cf_3.base west) + (-0.06cm, 0cm)$) {\scriptsize $\ldots$};
        
        \node[inner sep=1pt, anchor=base west] (wf_0) at ($(df_0.base west) + (-3.65cm, 0cm)$) {\scriptsize $y_1=$ \texttt{NNP}\ \ $x_i=$ Word: ``October''};
        \node[inner sep=1pt, anchor=base west] (wf_1) at ($(df_1.base west) + (-3.65cm, 0cm)$) {\scriptsize $y_1=$ \texttt{NNP}\ \ $x_i=$ Word: ``John''};
        \node[inner sep=1pt, anchor=base west] (wf_2) at ($(df_2.base west) + (-3.65cm, 0cm)$) {\scriptsize $y_1=$ \texttt{NNP}\ \ $x_i=$ Word: ``75th''};
        \node[inner sep=1pt, anchor=base west] (wf_3) at ($(df_3.base west) + (-3.65cm, 0cm)$) {\scriptsize $y_1=$ \texttt{NNP}\ \ $x_i=$ Word: ``two-tiered''};
        
        \node[repr, fill=darkgreen!70, minimum width=0.65cm, anchor=east] (p_0) at ($(wf_0.west) + (-0.12cm, -0.01cm)$) {};
        \node[repr, fill=darkgreen!70, minimum width=0.8cm, anchor=east] (p_1) at ($(wf_1.west) + (-0.12cm, -0.01cm)$) {};
        \node[repr, fill=darkgreen!70, minimum width=0.12cm, anchor=east] (p_2) at ($(wf_2.west) + (-0.12cm, -0.01cm)$) {};
        \node[repr, fill=darkgreen!70, minimum width=0.05cm, anchor=east] (p_3) at ($(wf_3.west) + (-0.12cm, -0.01cm)$) {};

        \begin{scope}[on background layer]
            \node [fill=lightgray!2, draw=black, rounded corners=2mm, thin, fit=(cf_0) (cf_3) (p_1)] (bg_recons) {};
        \end{scope}
        
        \node[inner sep=1pt] (vdots) at ($(bg_recons) + (0cm, 0.28cm)$) {\scriptsize $\vdots$};

        \draw[small_arrow] ($(bg_recons.south west) + (0.8cm, 0)$) to [out=-110, in=85] (p_label);

        \node[inner sep=1pt, anchor=north west] at ($(bg_recons.north west) + (0.05cm, -0.05cm)$) {\large \ding{173}};
        \node[inner sep=1pt, anchor=north west] at ($(bg_encoder.north west)!0.12!(bg_encoder.south west) + (0.02cm, 0cm)$) {\large \ding{172}};

    \end{tikzpicture}
    \caption{
      Model architecture of proposed model. \ding{172} is the ``CRF encoder w/ ELMo representations'' and \ding{173} is the ``reconstruction w/ hand-crafted features''.
    }
    \label{fig:framework}
  \end{figure}
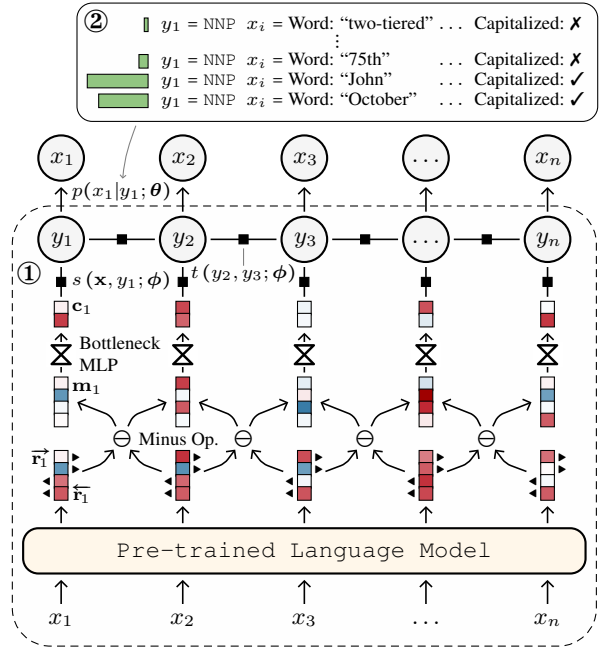
\section{Proposed Approach}

In this work, we for the first time propose a neural CRF-AE and leverage \textbf{PLM representations} and \textbf{hand-crafted features} for unsupervised POS tagging.

\subsection{CRF Encoder w/ PLM Representations}
\label{sec:crf-encoding}

As discussed in \S\ref{sec:vanilla-crf-ae}, the CRF-AE framework consists of two major components, i.e., the CRF encoder and the decoder for sentence reconstruction. 
We first introduce how to enhance the CRF encoder.
The major challenge of the CRF encoder is how to induce latent sequences more accurately via effective contextual representations.
Like most works before the DL era, \citet{ammar2014conditional} employ manually designed features to represent contexts. 

One of the major advances brought by DL is the strong capability of contextual representation via neural networks like LSTM and Transformer. Furthermore, pre-trained language models, such as ELMo and BERT, greatly amplify this advantage and are shown to be able to substantially improve performance for almost all NLP tasks. 

However, few works have tried to utilize such neural contextualized encoders for unsupervised POS tagging, except \citet{tran2016unsupervised} and \citet{gupta2020clustering}.
Most importantly, according to our knowledge, there is no work so far that successfully employ PLMs for unsupervised POS tagging. 

In this work, we propose to employ the contextual representations from PLM to enhance the CRF encoder of the CRF-AE model. 
Here we use ELMo \citep{peters-etal-2018-deep} to illustrate our method, which is the same for other PLMs  like BERT. 

\paragraph{ELMo outputs.} 
The encoder of ELMo consists of three layers \citep{peters-etal-2018-deep}. 
The bottom layer computes context-free word representations via word-wise character-level convolutional neural networks.
The top two layers, each with two unidirectional LSTMs (forward and backward), obtain context-aware word representations by concatenating the forward and backward representations.

After feeding an input sentence into ELMo, each word $x_i$ has three representation vectors, i.e., $(\mathbf{h}_i^0, \mathbf{h}_i^1, \mathbf{h}_i^2)$, corresponding to three encoder layers respectively. 
Following the standard practice, we take the weighted arithmetic mean (\texttt{ScalarMix}) of output vectors as the final contextualized word representation $\mathbf{r}_i$ for $\mathbf{x}_i$:
\begin{equation} \label{equ:scalarmix}
    \begin{split}
        \mathbf{r}_i =  \mathbf{\gamma}\sum_{k=0}^{K-1}\omega_k\mathbf{h}_i^k,
    \end{split}
\end{equation}
where $\omega_k ~~ (0 \le k < K)$ are softmax-normalized weights\footnote{The weights are trained only in the second stage of our training method.} and $K$ is the layer number; $\mathbf{\gamma }$ is the scale factor of the entire contextualized word representation. In our final model, we only use $\mathbf{h}_i^1$ and $\mathbf{h}_i^2$, since  including $\mathbf{h}_i^0$ degrades performance (see Table \ref{table:elmo_layer_result}). 

\paragraph{Minus operation.}
Apart from specific information of the focused word $x_i$, the contextualized word representation $\mathbf{r}_i$ from ELMo also contains a lot of common contextual information shared by neighbour words \citep{ethayarajh-2019-contextual}.
Therefore, inspired by previous works on constituent parsing \cite{wang-chang-2016-graph, cross-huang-2016-span}, 
we adopt the minus operation for representations as follows:
\begin{equation} \label{equ:minus-operation}
    \mathbf{m}_i=\left[\begin{matrix}\overrightarrow{\mathbf{r}}_{\!\!i}\\[5pt]\overleftarrow{\mathbf{r}}_{\!\!i}\end{matrix}\right]-\left[\begin{matrix}\overrightarrow{\mathbf{r}}_{\!\!i-1}\\[5pt]\overleftarrow{\mathbf{r}}_{\!\!i+1}\end{matrix}\right],
\end{equation}
where $\overrightarrow{\mathbf{r}}_{\!\!i}$ is the forward part of the final contextualized word representation $\mathbf{r}_i$ and $\overleftarrow{\mathbf{r}}_{\!\!i}$ is backward one. $\mathbf{m}_i$ is the word representation of $x_i$ after the minus operation.

\paragraph{Bottleneck MLP.} 
The ELMo adopts large dimensions $d$, i.e., $1024$, to encode as much information as possible.
Representations from ELMo contains syntax clues and even semantic ones besides the information about the POS.
Inspired by supervised dependency parsing models \citep{Timothy-d17-biaffine, li-eisner-2019-specializing}, we adopt a bottleneck MLP ($\mathtt{MLP}^\vertbowtie$), whose output vector has a very low dimension.
Because of the low dimension of the MLP output, redundant and irrelevant information will be stripped away:
\begin{equation}
    \begin{split}
        \mathbf{c}_i &= \mathtt{MLP}^\vertbowtie(\mathbf{m}_i) \\
                     &= \mathtt{LeakyReLU}\left(\mathbf{W}^\vertbowtie \cdot \mathtt{LayerNorm}(\mathbf{m}_i) + \mathbf{b}^\vertbowtie\right),
    \end{split}
\end{equation}
where the bottleneck size $d' \ll d$ is output dimensions of the bottleneck projection weight $\mathbf{W}^\vertbowtie \in \mathbb{R}^{d \times d'}$ and the bias $\mathbf{b}^\vertbowtie \in \mathbb{R}^{d'}$.

\paragraph{Scorer.}
The definition of a POS tagging sequence $\mathbf{y}$ given $\mathbf{x}$ is identical to equation \ref{equ:sequence-score}.
But the definition of bigram scores is different from the vanilla CRF-AE.
Here, a bigram score consists of two parts: a unigram score $s\left(\mathbf{x}, y_i; \boldsymbol{\phi}\right)$ estimated from ELMo representations and a matrix-maintained transition score $t\left(y_{i-1}, y_i; \boldsymbol{\phi}\right)$.
\begin{equation}
        s\left(\mathbf{x}, y_{i-1}, y_i; \boldsymbol{\phi}\right) = s\left(\mathbf{x}, y_i; \boldsymbol{\phi}\right) + t\left(y_{i-1}, y_i; \boldsymbol{\phi}\right).
\end{equation}
Specifically, $s\left(\mathbf{x}, y_i\right)$ is calculated as follows:
\begin{equation}
    s\left(\mathbf{x}, y_i; \boldsymbol{\phi}\right) = \mathtt{LayerNorm}(\mathbf{W}^s \cdot \mathbf{c}_i + \mathbf{b}^s)\left[y_i\right],
\end{equation}
where $\mathbf{W}^s \in \mathbb{R}^{d'\times \left|\mathcal{Y}\right|}$ is the projection weight of scoring, $\mathbf{b}^s \in \mathbb{R}^{\left|\mathcal{Y}\right|}$ is the scoring bias, and $\mathcal{Y}$ is the POS tag set. $\left[y_i\right]$ is the index selection operation.

\subsection{Reconstruction w/ Hand-crafted Features}
\begin{table}[tb]
    \centering
    \renewcommand{\arraystretch}{1.2}
    \begin{tabular}{l|*{2}{p{2.4em}}p{4.1em}}
    \toprule
        \rowcolor[gray]{.9}
        Feature                  & John           & 75th         & two-tiered \\ 
    \midrule 
        Word                     & John           & 0th$^\dagger$ & \texttt{UNK}$^\ddagger$   \\ 
        \rowcolor[gray]{.95}
        {Uni-gram Suffix\,}          & n              & h             & d              \\
        Bi-gram Suffix           & hn             & th            & ed             \\
        \rowcolor[gray]{.95}
        Tri-gram Suffix          & ohn            & 0th           & red            \\
        Has Digit                & \ding{55}      & \ding{51}     & \ding{55}      \\
        \rowcolor[gray]{.95}
        Has Hyphen               & \ding{55}      & \ding{55}     & \ding{51}      \\
        Capitalized              & \ding{51}      & \ding{55}     & \ding{55}      \\
    \bottomrule
    \end{tabular}
    \caption{Feature templates for feature-rich reconstruction. $\dagger$: before extracting features, we replace continuous digits into a single ``0'' in each word. $\ddagger$: features appeared less then 50 times in the training data are replaced with a special \texttt{UNK} feature.}
    \label{table:feature_example}
\end{table}
In \citet{ammar2014conditional}, the reconstruction probabilities are stored and updated as a matrix. 
The conditional probability $p(x_i\mid y_i)$, i.e., generating $x_i$ given $y_i$, is modeled at the whole-word level.
This leads to the data sparseness problem.
For rare words, the probabilities are usually unreliable.

Therefore, we borrow the idea of feature-rich HMM by  \citet{berg2010painless}. 
The idea is to utilize rich morphological  information to learn more reliable generation probability. 
For example, suffixes usually provide strong clues to POS categories. 
In this work, we adopt the feature templates proposed by \citet{berg2010painless}, as shown in Table~\ref{table:feature_example}.

With the hand-crafted features, we then parameterize tag-to-word emission probabilities as local multinomials: 
\begin{equation} \label{equ:emit-scores}
    p(x_i\mid y_i; \boldsymbol{\theta})=\frac{\exp{\left(\boldsymbol{\theta} \cdot f\left(x_i, y_i\right)\right)}}{\sum_{x' \in \mathcal{V}}{\exp{\left(\boldsymbol{\theta}\cdot f\left(x', y_i\right)\right)}}}
\end{equation}
where $\boldsymbol{\theta}$ is the feature weight vector and $\mathcal{V}$ is the vocabulary set.

\section{Experiments on English PTB}

\subsection{Settings}\label{subsec:exp-settings}

\paragraph{Data.} 
\label{subsec:data} 
Following previous works on unsupervised POS tagging, we conduct experiments on the Wall Street Journal (WSJ) data from PTB, yet with two distinct data settings.

\begin{asparaenum}[(1)]
\item \emph{WSJ-All}. Almost all previous works train and evaluate their models on the entire WSJ data. We report results on WSJ-All for comparison with previous works. However, this data setting is very unfriendly for selecting hyper-parameters, such as stopping and best epoch numbers, M-1 mappings, learning rates, network dimensions, etc. It is probable that some previous works make modeling choices by directly looking at the evaluation performance, since training loss (e.g., data likelihood) is quite loosely correlated with performance. Such details are usually omitted or only implicitly discussed in previous works. 

\item \emph{WSJ-Split}. We follow the practice in  unsupervised dependency parsing and divide the WSJ dataset into train (sections 02-21), dev (section 22) and test (section 23). 
We tune hyper-parameters and study the contributions of individual model components by referring to performance on WSJ-Dev. Moreover, we determine the best many-to-one mappings on WSJ-Dev, which are directly used to compute many-to-one accuracy (M-1) on both WSJ-Dev and WSJ-Test. 
\end{asparaenum}

\textbf{We strongly suggest that in future researchers can adopt the WSJ-Split setting}. 
First, the WSJ-Split setting is more realistic because it is able to evaluate a model's generalization ability with  out-of-vocabulary words. 
Second, it is more reasonable and fairer to use WSJ-dev to choose hyper-parameters and it is usually feasible to manually annotate a dev data, even if very small-scale.

\paragraph{Evaluation metrics.} 
Following previous works, we mainly adopt many-to-one accuracy,  %
and also report one-to-one accuracy (1-1) and validity-measure (VM) values for better comparison.  
To reduce the effect of performance vibration, we follow previous works, run each model for five times with different seeds, and report the mean and standard deviation.
Please see Appendix~\ref{sec:details_of_metric} for details. 

\paragraph{Hyper-parameters.} 
We set the number of predicted POS tags to $45$ and the output dimensions of $\mathtt{MLP}^\vertbowtie$ to $5$.
We train each model on the training data for at most $50$ epochs, and select the best epoch based on data log-likelihood (LL). 
Please see Appendix~\ref{sec:details_of_hyper_parameters} for full details of hyper-parameters.

\paragraph{Three-stage Training procedure.}
\label{para:training_proc}
Unsupervised models are very sensitive to parameter initialization. %
Inspired by previous works \citep{han-etal-2017-dependency, he2018unsupervised}, we adopt a three-step progressive training procedure. 
\begin{inparaenum}[1)]
\item %
We train a feature-rich HMM model from random initialization, 
and produce the 1-best prediction from it for each training sentence. 
\item %
The feature-rich HMM model is used as a teacher to pre-train the CRF-AE model. More concretely, 
we train the CRF encoder on the pseudo-labeled training data in a supervised fashion for $5$ epochs; meanwhile we directly copy the feature weights from the feature-rich HMM model to the decoder of the CRF-AE model. 
\item %
We train our full CRF-AE model on unlabeled training data %
with parameters obtained in the second step as initialization. %
\end{inparaenum}

\subsection{Model Development on WSJ-Split}

\begin{table}[tb]
    \setlength{\tabcolsep}{1.8pt}
    \renewcommand{\arraystretch}{1.35}
    \centering
    \begin{tabular}{p{4.6em}|cccc}
    \toprule
        \rowcolor[gray]{.9}
        Layer                        & M-1                             & 1-1                             & VM                              & LL \\ 
    \midrule 
        \namebox{\phantom{\{}0}      & \resultbox{79.98}{0.3}          & \resultbox{59.13}{3.4}          & \resultbox{73.06}{0.9}          & \resultbox{-73.06}{1.4} \\
        \rowcolor[gray]{.95}
        \namebox{\phantom{\{}1}      & \resultbox{82.61}{0.8}          & \resultbox{63.03}{5.0}          & \resultbox{76.98}{1.2}          & \resultbox{-79.52}{0.8} \\
        \namebox{\phantom{\{}2}      & \resultbox{82.45}{1.0}          & \resultbox{60.29}{3.2}          & \resultbox{76.27}{1.0}          & \resultbox{-83.35}{0.5} \\ 
        \hline
        \rowcolor[gray]{.95}
        \namebox{\{0, 1, 2\}}\topsh  & \resultbox{81.53}{0.3}          & \resultbox{64.03}{4.1}          & \resultbox{76.21}{0.7}          & \resultbox{-76.89}{0.4} \\ 
        \namebox{\{1, 2\}$^\oplus$}  & \resultbox{82.28}{1.3}          & \resultbox{63.54}{4.5}          & \resultbox{76.91}{1.3}          & \resultbox{-78.96}{0.4} \\ 
        \rowcolor[gray]{.95}
        \namebox{\{1, 2\}}           & \resultbox{\textbf{83.20}}{0.7} & \resultbox{\textbf{65.17}}{2.3} & \resultbox{\textbf{77.69}}{0.7} & \resultbox{-80.49}{0.5} \\ 
    \bottomrule
    \end{tabular}
    \caption{
    Results of utilizing different layers of the ELMo on  {WSJ-Dev}. $\bigoplus$ means directly concatenating the representation vectors of different layers. 
    }
    \label{table:elmo_layer_result}
\end{table}

\paragraph{Using which ELMo layers.}\label{subsec:layer_exp}

As mentioned above, ELMo produces three representation vectors for each word $x$, corresponding to its three encoder layers. 
Since the usefulness of information contained in different ELMo layers is unknown for our task at hand, we conduct experiments to study which layers to use and how to use them. 
Table \ref{table:elmo_layer_result} shows the results.
When using single-layer representations, 
it is obvious that using one of the top two layers (1/2) is superior to using the bottom $0$-th layer\footnote{Minus operations do not apply to vectors at the 0-th layer, i.e., context-independent word type embeddings, which are directly used as $\mathbf{m}_i$.}. This is in line with our expectation considering that the $0$-th layer corresponds to context-independent word type embeddings. 
The first layer is superior to the second one, which is consistent with \citet{peters-etal-2018-deep}, who also conclude that {the information contained by the first layer are more suitable for POS tagging than the second layer.}

Then we try to combine multiple layers by 
using aforementioned \texttt{ScalarMix} in Equation~\ref{equ:scalarmix}. 
It is clear that using the top two contextualized layers (\{1, 2\}) achieves best performance. 
We find that the weight contribution of layer 1 and 2 is about 92\% vs. 8\%, confirming again that the first contextualized layer provides the majority of syntactic information, while the second layer is more concerned with high-level semantics. 
We can also see that replacing \texttt{ScalarMix} with simple concatenation leads to large performance drop.

Comparing M-1, 1-1, VM, and LL, we can see that M-1, 1-1 and VM are highly correlated, whereas LL is quite loosely correlated with model performance, suggesting that training loss cannot be used for selecting models or tuning hyper-parameters.

\vspace{+1.75ex}

In the following, we try to understand \textbf{the contribution of different components} by removing one from the full CRF-AE model at a time. Table \ref{table:ablation_result} shows the results. 

\begin{table}[tb]
    \renewcommand{\arraystretch}{1.35}
    \setlength{\tabcolsep}{1.9pt}
    \centering
    \begin{tabular}{l|ccc}
    \toprule
        \rowcolor[gray]{.9}
        Model                          &  M-1                            & 1-1                             &  \multicolumn{1}{c}{VM}         \\
    \midrule 
        Full CRF-AE          & \resultbox{\textbf{83.20}}{0.7} & \resultbox{\textbf{65.17}}{2.3} & \resultbox{\textbf{77.69}}{0.7} \\
        \rowcolor[gray]{.95}
        \quad w/o Features   & \resultbox{76.74}{1.4}          & \resultbox{61.34}{3.5}          & \resultbox{73.55}{1.0}          \\
        {\quad w/o PLM Repr.\quad}  & \resultbox{78.40}{0.9}          & \resultbox{61.31}{4.9}          & \resultbox{72.55}{1.6}          \\
        \rowcolor[gray]{.95}
        \quad w/o Minus Op.  & \resultbox{81.28}{1.5}          & \resultbox{63.07}{2.8}          & \resultbox{76.04}{1.1}          \\
        \quad w/o 3-stage Train  & \resultbox{80.21}{3.4}          & \resultbox{59.50}{2.9}          & \resultbox{75.70}{1.2}          \\
        \rowcolor[gray]{.95}
        ELMo $\to$ BERT      & \resultbox{82.30}{1.0}          & \resultbox{62.78}{5.8}          & \resultbox{76.13}{1.6}          \\
    \bottomrule
    \end{tabular}
    \caption{
    The contribution of different components on {WSJ-Dev} by removing one component at a time. 
    }
    \label{table:ablation_result}
\end{table}

\paragraph{Usefulness of hand-crafted features.} 
In order to measure the effectiveness of hand-crafted features in the reconstruction part, 
we revert to the vanilla matrix-maintained version.
We can see that rich hand-crafted features are critical and not using them leads to the largest performance drop. 

\paragraph{Usefulness of PLMs.} 
We first replace pre-trained ELMo with a conventional three-layer BiLSTM encoder that is trained from scratch. 
We use pre-trained word embeddings of  \citet{he2018unsupervised} as encoder inputs. 
As expected, performance also declines a lot. 
It shows that ELMo does provide very useful information. 
{
We have also tried to replace ELMo with BERT without much hyper-parameter tuning, as shown in the bottom row, but found that the performance decreases. The results are similar on the multilingual UD data in Table \ref{table:multilingual_new_settings_mo}. 
We suspect the reasons are two-fold. First, we did not carefully tune the hyper-parameters for using BERT due to time and resource limitation. 
Second, we suspect the ELMo word representations suffice and are even more suitable for unsupervised POS tagging. The POS tag of a word usually heavily depends on neighboring words within a small window, which makes the BiLSTM encoder superior to Transformer. The latter is more powerful to capture long-distance dependencies.  
}

\paragraph{Usefulness of minus operation.} 
Besides the minus operation in Equation~\ref{equ:minus-operation}, the default choice is directly using the ELMo output, i.e., $\mathbf{r}_i$.
As shown in the fourth row, models without the minus operation are inferior to the models with the minus operation. 
We believe it is because the original ELMo representations have a lot of common contextual information shared by neighbour words, and the minus operation can remove them.

\paragraph{Usefulness of the three-stage training procedure.}
To find out the effect of our three-step progressive training procedure, we randomly initialize model parameters. 
The result shows that the randomly initialization decreases model performances substantially. 
It proves that the three-stage training procedure helps models find relatively good initial parameters.

\begin{table}[tb]
    \renewcommand{\arraystretch}{1.35}
    \setlength{\tabcolsep}{0.8pt}
    \centering
    \begin{tabular}{l|ccc}
    \toprule
        \rowcolor[gray]{.9}
        Model                       &  M-1                            & 1-1                             & VM \\
    \midrule 
        \namebox{HMM (re-Impl.)}    & \resultbox{65.25}{2.0}          & \resultbox{47.62}{2.6}          & \resultbox{58.16}{1.0}          \\
        \rowcolor[gray]{.95}
        \namebox{FHMM (re-Impl.)}   & \resultbox{73.91}{1.0}          & \resultbox{59.13}{6.4}          & \resultbox{68.69}{2.5}          \\
        \namebox{INP-GHMM (re-Run)\quad} & \resultbox{76.10}{1.5}          & \resultbox{53.83}{3.2}          & \resultbox{72.19}{0.8}          \\
    \hline
        \rowcolor[gray]{.95}
        \namebox{Ours}\topsh              & \resultbox{\textbf{82.89}}{0.7} & \resultbox{\textbf{65.32}}{2.5} & \resultbox{\textbf{78.06}}{0.8} \\
    \bottomrule
    \end{tabular}
    \caption{
    Results on {WSJ-Test}. We re-implement HMM and Feature-rich HMM and re-run INP-GHMM.
    }
    \label{table:new_settings_result}
\end{table}

\subsection{Results on WSJ-Test}

We report results on {WSJ-Test} in  Table~\ref{table:new_settings_result} and hope future researchers adopt the WSJ-Split setting. Considering that INP-GHMM is the current SOTA model on English PTB, we re-run their open-source code\footnote{https://github.com/jxhe/struct-learning-with-flow} with default configuration on WSJ-Split.  
We re-implement vanilla HMM and feature-rich HMM of \citet{berg2010painless}, and train them with Adam algorithm via direct gradient descent.  %
Results show that our model is superior to the previous best one, and achieves current SOTA results.

\subsection{Performance Comparison on WSJ-All}
In order to compare with previous works, we report results on WSJ-All in Table \ref{table:main_result}.
We directly use all hyper-parameters obtained from WSJ-Split.  %

We can see that our proposed model outperforms all previous works by large margin. 
The INP-GHMM model \citep{he2018unsupervised} achieves the previous best performance on WSJ-All. 
Our model outperforms theirs by $2.41$ and $3.54$ on M-1 and VM, respectively.

\section{Experiments on Multilingual UD}
\subsection{Settings}
\label{subsec:ud_settings}
\paragraph{Data.} 
For more thorough comparison with previous works, we also report results\footnote{We run each model for five times with different random seeds.} on the Multilingual Universal Dependencies treebank v2.0 (UD), consisting of 10 languages \citep{mcdonald-etal-2013-universal}.
Similar to experiments on English PTB, we adopt two settings for the UD data, i.e., UD-Split and UD-All. For UD-Split, we adopt the default partition of the UD data.

\begin{table}[tb]
    \setlength{\tabcolsep}{1.2pt}
    \renewcommand{\arraystretch}{1.35}
    \centering
    \begin{tabular}{l|ccc}
    \toprule
        \rowcolor[gray]{.9}
        Model                      & \multicolumn{1}{c}{M-1}          & \multicolumn{1}{c}{1-1}          & \multicolumn{1}{c}{VM}          \\ 
    \midrule
        \namebox{HMM (B'10)}       & \resultbox{63.1}{1.3}         & \namebox{--}                     & \namebox{--}                    \\ 
        \rowcolor[gray]{.95}
        \namebox{FHMM (B'10)}      & \resultbox{75.5}{1.1}         & \namebox{--}                     & \namebox{--}                    \\ 
        \namebox{FHMM (re-Impl.)}  & \resultbox{74.70}{2.2}           & \resultbox{60.88}{4.3}           & \resultbox{68.53}{2.1}          \\ 
        \rowcolor[gray]{.95}
        \namebox{Brown (C'10)}     & \namebox{76.1}                & \namebox{60.7}                & \namebox{68.8}               \\ 
        \namebox{S-CODE (Y'12)}    & \resultbox{80.23}{0.7}           & \namebox{--}                     & \resultbox{72.07}{0.4}          \\ 
        \rowcolor[gray]{.95}
        \namebox{GHMM (L'15)}      & \resultbox{75.4}{1.0}         & \namebox{--}                     & \resultbox{68.5}{0.5}        \\ 
        \namebox{NHMM (T'16)}      & \namebox{79.1}                & \namebox{60.7}                & \namebox{71.7}               \\ 
        \rowcolor[gray]{.95}
        \namebox{INP-GHMM (H'18)\quad}  & \resultbox{80.8}{1.3}         & \namebox{--}                     & \resultbox{74.1}{0.7}        \\ 
        \namebox{MIM (S'19)}       & \resultbox{78.1}{0.8}         & \namebox{--}                     & \namebox{--}                    \\ 
        \rowcolor[gray]{.95}
        \namebox{SyntDEC (G'20)}   & \resultbox{78.2}{0.9}         & \namebox{--}                     & \namebox{--}                    \\ 
    \hline
        \namebox{Ours}\topsh       & \resultbox{\textbf{83.21}}{1.2}  & \resultbox{\textbf{65.78}}{2.8}  & \resultbox{\textbf{77.64}}{0.5} \\ 
    \bottomrule
    \end{tabular}
    \caption{
    Results on {WSJ-All}.
    Here, B'10 is for \citet{berg2010painless}, 
    C'10 for \citet{christodoulopoulos-etal-2010-two}, 
    Y'12 for \citet{yatbaz-etal-2012-learning}, 
    L'15 for \citet{lin2015unsupervised}, 
    T'16 for \citet{tran2016unsupervised}, 
    H'18 for \citet{he2018unsupervised}, 
    S'19 for \citet{stratos2019mutual}, 
    and, G'20 for \citet{gupta2020clustering}.}
    \label{table:main_result}
\end{table}
\begin{table*}[tb]
    \centering
    \renewcommand{\arraystretch}{1.35}
    \setlength{\tabcolsep}{0.35pt}
    \begin{tabular}{l|*{11}c}
        \toprule
        \rowcolor[gray]{.9}
        UD-Dev                                   & de$^\ast$                         & en$^\ast$                         & es$^\ast$                         & fr$^\ast$                         & id                                & it$^\ast$                         & ja                              & ko                              & pt-br$^\ast$                    & sv                              & Mean           \\
        \midrule
        \namebox{Full CRF-AE}                    & \resultboxud{76.18}{4.0}          & \resultboxud{\textbf{80.30}}{2.2} & \resultboxud{\textbf{81.76}}{1.0} & \resultboxud{\textbf{82.56}}{0.4} & \resultboxud{\textbf{80.99}}{0.5} & \resultboxud{\textbf{80.32}}{1.8} & \resultboxud{86.61}{1.1}          & \resultboxud{73.00}{3.0}          & \resultboxud{\textbf{81.38}}{2.2} & \resultboxud{\textbf{74.93}}{2.3} & \namebox{\textbf{79.80}} \\
        \rowcolor[gray]{.95}
        \namebox{\quad w/o Features}             & \resultboxud{70.18}{1.2}          & \resultboxud{73.64}{1.3}          & \resultboxud{73.91}{6.0}          & \resultboxud{72.51}{2.8}          & \resultboxud{73.09}{2.3}          & \resultboxud{68.18}{1.3}          & \resultboxud{75.96}{3.8}          & \resultboxud{63.36}{3.3}          & \resultboxud{68.95}{6.0}          & \resultboxud{65.81}{4.4}          & \namebox{70.56} \\
        \namebox{\quad w/o UD Adjust.}           & \resultboxud{73.94}{1.8}          & \resultboxud{73.63}{4.0}          & \resultboxud{77.95}{3.1}          & \resultboxud{76.05}{3.1}          & \resultboxud{76.57}{1.4}          & \resultboxud{72.12}{4.7}          & \resultboxud{82.19}{1.1}          & \resultboxud{74.37}{3.2}          & \resultboxud{74.47}{3.2}          & \resultboxud{64.96}{3.3}          & \namebox{74.62} \\
        \rowcolor[gray]{.95}
        \namebox{\quad w/o Language Adjust.\ \ } & \resultboxud{75.99}{1.2}          & \resultboxud{78.97}{2.3}          & \resultboxud{79.66}{2.1}          & \resultboxud{79.60}{2.0}          & \namebox{$\Leftrightline$}        & \resultboxud{71.71}{6.5}          & \namebox{$\Leftrightline$}      & \namebox{$\Leftrightline$}      & \resultboxud{73.72}{1.9}          & \namebox{$\Leftrightline$}      & \namebox{77.52} \\
        \namebox{\quad w/o PLM Repr.}            & \resultboxud{75.11}{2.7}          & \resultboxud{76.50}{1.4}          & \resultboxud{78.78}{0.9}          & \resultboxud{82.16}{1.7}          & \resultboxud{77.96}{1.5}          & \resultboxud{70.54}{2.6}          & \resultboxud{82.26}{1.0}          & \resultboxud{65.47}{1.4}          & \resultboxud{79.11}{2.6}          & \resultboxud{68.94}{1.7}          & \namebox{75.68} \\
        \rowcolor[gray]{.95}
        \namebox{\quad w/o 3-stage Training}     & \resultboxud{\textbf{77.52}}{3.1} & \resultboxud{72.18}{3.6}          & \resultboxud{74.70}{3.8}          & \resultboxud{78.26}{3.2}          & \resultboxud{78.62}{2.8}          & \resultboxud{70.85}{2.2}          & \resultboxud{83.93}{1.4}          & \resultboxud{\textbf{76.01}}{2.7} & \resultboxud{77.26}{5.7}          & \resultboxud{68.50}{5.5}          & \namebox{75.78} \\
        \namebox{ELMo $\to$ mBERT}               & \resultboxud{75.96}{4.2}          & \resultboxud{78.12}{1.3}          & \resultboxud{79.67}{1.6}          & \resultboxud{81.09}{0.7}          & \resultboxud{80.13}{0.5}          & \resultboxud{75.66}{3.0}          & \resultboxud{\textbf{86.92}}{1.5} & \resultboxud{73.18}{3.1}          & \resultboxud{80.69}{2.4}          & \resultboxud{72.87}{2.8}          & \namebox{78.43} \\
        \midrule 
        \midrule
        \rowcolor[gray]{.9}
        UD-Test                                  & de                       & en                       & es                       & fr                       & id                              & it                       & ja                              & ko                              & pt-br                    & sv                              & Mean           \\
        \midrule
        \namebox{HMM (re-Impl.)}                 & \resultboxud{60.13}{1.2}          & \resultboxud{63.85}{2.4}          & \resultboxud{64.68}{3.8}          & \resultboxud{65.50}{4.5}          & \resultboxud{66.23}{2.1}          & \resultboxud{66.14}{1.8}          & \resultboxud{60.02}{0.4}          & \resultboxud{46.55}{0.6}          & \resultboxud{57.65}{6.3}          & \resultboxud{57.07}{5.1}          & \namebox{60.78} \\
        \rowcolor[gray]{.95}
        \namebox{FHMM (re-Impl.)}                & \resultboxud{70.95}{2.7}          & \resultboxud{75.58}{0.7}          & \resultboxud{76.26}{1.2}          & \resultboxud{77.33}{1.8}          & \resultboxud{73.67}{1.0}          & \resultboxud{74.73}{2.5}          & \resultboxud{72.47}{0.5}          & \resultboxud{63.77}{1.7}          & \resultboxud{77.67}{2.1}          & \resultboxud{67.99}{2.3}          & \namebox{73.04} \\
        \namebox{GHMM (re-Run)}                  & \resultboxud{81.95}{1.2}          & \resultboxud{75.49}{1.5}          & \resultboxud{78.92}{1.7}          & \resultboxud{73.48}{7.4}          & \resultboxud{76.09}{4.3}          & \resultboxud{72.87}{4.5}          & \resultboxud{75.41}{1.1}          & \resultboxud{68.31}{1.7}          & \resultboxud{74.84}{5.5}          & \resultboxud{72.15}{3.6}          & \namebox{74.95} \\
        \rowcolor[gray]{.95}
        \namebox{INP-GHMM (re-Run)}              & \resultboxud{82.79}{1.1}          & \resultboxud{75.93}{1.5}          & \resultboxud{79.61}{2.9}          & \resultboxud{73.55}{7.2}          & \resultboxud{76.92}{3.6}          & \resultboxud{73.60}{4.8}          & \resultboxud{76.32}{1.2}          & \resultboxud{67.85}{2.3}          & \resultboxud{75.43}{5.0}          & \resultboxud{74.33}{3.5}          & \namebox{75.63} \\
        \hline
        \namebox{Ours}\topsh                     & \resultboxud{77.46}{4.5}          & \resultboxud{\textbf{79.60}}{2.2} & \resultboxud{\textbf{80.46}}{0.9} & \resultboxud{\textbf{79.36}}{0.6} & \resultboxud{\textbf{80.77}}{0.5} & \resultboxud{\textbf{80.82}}{2.2} & \resultboxud{\textbf{79.93}}{2.5} & \resultboxud{\textbf{75.48}}{3.1} & \resultboxud{\textbf{81.23}}{2.3} & \resultboxud{\textbf{76.29}}{2.0} & \namebox{\textbf{79.14}} \\
        \rowcolor[gray]{.95}
        \namebox{Ours (GHMM Init.)}              & \resultboxud{\textbf{84.77}}{2.2} & \namebox{--}                      & \namebox{--}                      & \namebox{--}                      & \namebox{--}                      & \namebox{--}                      & \namebox{--}                      & \namebox{--}                      & \namebox{--}                      & \namebox{--}                      & \namebox{--} \\
        \bottomrule
    \end{tabular}
    \caption{M-1 accuracy on UD-Split. \textbf{Upper Part:} The contribution of different components on UD-Dev by removing one component at a time. $\ast$ means adopting the language-specific suffix features for this language. ``$\Leftrightline$'' means the result is identical to that of Full CRF-AE. \textbf{Lower Part:} Performance comparison on UD-Test. %
    }
    \label{table:multilingual_new_settings_mo}
\end{table*}

\paragraph{Hyper-parameters.} We directly adopt most hyper-parameters obtained for PTB with three important exceptions. 
First, The number of predicted POS tags is changed to $12$.
Second, since the scale of data for each language diverge a lot, we adjust the feature cutoff threshold to be proportional to the token number against English partition. For example, the ``de'' data contains about 293k
 tokens, which is about 28\% of that of ``en'' (1M), and therefore we set the threshold to 14 ($28\% \times 50$). 
Third, we adjust the hand-crafted features to accommodate the 12-tag UD standard and characteristics of different languages, detailed in the following. 

\paragraph{Modifications on hand-crafted features.}
\label{para:hand_crafted_features_ud}
The fine-grained 45-tag WSJ standard is greatly different from the coarse-grained 12-tag UD standard adopted by the multilingual UD datasets \citep{petrov-etal-2012-universal}. 
Therefore, we start from the features of  \citet{berg2010painless} in Table \ref{table:feature_example} as the base, and make adjustments from two aspects. 

\begin{asparaenum}[(1)]
\item \emph{Adjustments for UD}. 
We remove the ``Capitalized'' feature, which is originally purposed to distinguish proper and common nouns which correspond to a single UD tag. 
Moreover, we replace all punctuation marks with a special ``\texttt{PUNCT}'' word form, add a new feature template ``is-Punctuation'', as UD uses a single tag for punctuation marks. 

\item \emph{Adjustments for specific languages\footnote{We only adopt language-specific adjustments for ``de'', ``en'', ``es'', ``fr'', ``it'' and ``pt-br''.}.} 
The UD tag set doesn't distinguish inflections such as numbers, tenses, and genders.
We find this can be accommodated by customizing suffix uni/bi/tri-gram features. 
We simply remove a certain number of ending characters (related to inflectional affixes) for a word form before extracting suffix features. 
We remove the last character for ``it'', and the last two characters for ``de''.
For ``fr'', ``es'', and ``pt-br'', we remove last two characters if the word ends with ``s'', and the last one otherwise. 
For ``en'', we only remove the last ``s'' letter if applicable. \end{asparaenum}

\subsection{Results on UD-Split}
\label{subsec:results_on_ud_split}

Table~\ref{table:multilingual_new_settings_mo} shows the M-1 results. For 1-1 and VM-results, please refer to Table~\ref{table:multilingual_new_settings_oo} and Table~\ref{table:multilingual_new_settings_vm} in the Appendix. 

We perform ablation study on UD-Dev. 
Most of the results show the same trend as on WSJ-Dev. 
In particular, we find that our two adjustment strategies for the UD data are very helpful, and the UD adjustment is more helpful. 
After observation, we find that without UD-specific adjustments, punctuation marks are more likely to be divided into multiple tags. For example, models may assign three different tags to periods, commas, and quotation marks. 
Moreover, with the removal of the ``Capitalized'' feature, which is one of the UD adjustments, the models 
no longer distinguish common and proper nouns and assign one tag to them.\footnote{However, we find that some models still divide nouns into multiple tags by some unknown criteria.}

Without language-specific adjustments, highly inflected languages, e.g., Italian (it) and Brazilian Portuguese (pt-br), are more likely to distinguish words by their number or gender rather than part-of-speech.
For example, in English, models without making language-specific adjustments will tend to split nouns into two classes: single nouns and plural nouns ending with ``s''.

We report the M-1 results on UD-Test in Table~\ref{table:multilingual_new_settings_mo}.
We run our implemented vanilla HMM and feature-rich HMM, and the latter adopt the same features after UD and linguage adjustments. 
Unfortunately, we are unable to re-run SyntDEC, the current SOTA on UD-All, since its authors \citep{gupta2020clustering} have not yet released their code. 
We also re-run INP-GHMM \citep{he2018unsupervised} with their released code, which is the current SOTA on WSJ-All. 
We take context-free word representations (0-th layer) of ELMo as inputs of INP-GHMM, which should be better than Skip-Gram embeddings. %
Please see Appendix~\ref{sec:details_of_inp_ghmm_hyper_parameters} for details of hyper-parameters.

Results show that our models achieve the highest M-1 accuracy on $9$ out of $10$ languages, except ``de''.
After investigation on why our models fail to outperform INP-GHMM on ``de'', we find that the direct reason is that INP-GHMM is initialized with GHMM, and the simple GHMM is already more superior to our model. 
Therefore, we replace FHMM with GHMM in the first stage of our training procedure. 
Results show that %
our models are substantially improved in ``de''. 
However, we still do not understand the reason behind these results, which we leave for future investigation due 
to time limitation.

\subsection{Performance Comparison on UD-All}
To compare with previous works, we report results on UD-All in Table~\ref{table:multilingual_result} in the Appendix. 
For thorough comparison, we also re-run GHMM and INP-GHMM on UD-All. 
The results show identical trends as those on UD-Split.  
\section{Related Works}

\paragraph{Unsupervised POS tagging.} 
In addition to HMMs and the CRF-AE, other approaches for unsupervised POS tagging are as follows.

\begin{asparaenum}[(1)]
\item \emph{Clustering.} 
The clustering approach, as a mainstream unsupervised learning technique, is also investigated for unsupervised POS tagging \citet{yatbaz-etal-2012-learning,yuret-etal-2014-unsupervised, gupta2020clustering}.
All these works adopt the \textit{k}-means algorithm to divide word tokens into different groups. 
The main difference among them is how to represent words. 
\citet{yatbaz-etal-2012-learning} propose to learn context-free word embeddings by minimizing the distance between each word and its substituted words. Substituted words are selected according to a n-gram language model. 
\citet{yuret-etal-2014-unsupervised} extend their previous work to produce context-sensitive word embeddings. 
\citet{gupta2020clustering} adopt a deep clustering approach that uses a feed-forward neural network to transform word representations from mBERT into a lower-dimension clustering-friendly space. 
Transformation with reconstruction loss and clustering are jointly trained. 
Unfortunately, all three works have not released their source code.

\item \emph{Mutual information maximization.} 
The mutual information maximization approach is proposed by \citet{stratos2019mutual}.
The idea is that we can predict POS tags in two ways (using the words themselves or their context), and predictions from these two ways should agree as more as possible.
\end{asparaenum}

\paragraph{Utilizing PLMs for unsupervised tagging or parsing.} 
As discussed earlier, SyntDEC \citep{gupta2020clustering} is the only work that employs PLMs for unsupervised POS tagging based on deep clustering. 
As for unsupervised parsing, \citet{wu-etal-2020-perturbed} propose a perturbed masking technique to estimate inter-word correlations and then induce syntax trees from those correlations.
\citet{kim-etal-2020-are} extract constituency trees from the PLMs through capturing syntactical proximity between representations of two adjacent words (or subwords). 
If the proximity is loose, then it is likely that the middle position of the two words corresponds to some constituent boundary. 
\citet{cao-etal-2020-unsupervised} successfully exploit PLMs for unsupervised constituency parsing based on constituency test, achieving SOTA performance.

\paragraph{Utilizing CRF-AE.} 
\citet{cai-etal-2017-crf} apply CRF-AE to unsupervised dependency parsing. 
They use the encoder to generate a most likely dependency tree and then force the decoder to reconstruct the input sentence from the tree.
\citet{zhang-etal-2017-semi} propose a neural CRF-AE for semi-supervised learning on sequence labeling problems (including POS tagging) and \citet{jia-etal-2020-semi} adopt a neural CRF-AE for semi-supervised semantic parsing.

\section{Conclusions}
This work bridges PLMs and hand-crafted features for unsupervised POS  tagging. 
Based on the CRF-AE framework, we employ powerful contextualized representations from PLMs in the CRF encoder, and incorporate rich morphological features for better reconstruction. 
Our proposed approach achieves new SOTA on 45-tag English PTB and 12-tag multilingual UD datasets, outperforming previous results by large margin. 
Experiments and analysis show that rich features and PLM representations are critical for the superior performance of our model. 
Meanwhile, simple adjustments of hand-crafted features are key for the success of our model on languages other than English. 

\section*{Acknowledgments}
We thank the anonymous reviewers for the helpful comments.
We are very grateful to Wei Jiang for his early-stage exploration on  unsupervised POS tagging.
We also thank Chen Gong, Yu Zhang, Ying Li, Qingrong Xia, Yahui Liu, and Tong Zhu for their help in paper writing and polishing. This work was supported by National Natural Science Foundation of China (Grant No. 62176173, 61876116) and a Project Funded by the Priority Academic Program Development (PAPD) of Jiangsu Higher Education Institutions.

\bibliography{acl22}

\begin{thebibliography}{39}
\expandafter\ifx\csname natexlab\endcsname\relax\def\natexlab#1{#1}\fi

\bibitem[{Ammar et~al.(2014)Ammar, Dyer, and Smith}]{ammar2014conditional}
Waleed Ammar, Chris Dyer, and Noah~A. Smith. 2014.
\newblock \href
  {https://proceedings.neurips.cc/paper/2014/hash/b9f94c77652c9a76fc8a442748cd54bd-Abstract.html}
  {Conditional random field autoencoders for unsupervised structured
  prediction}.
\newblock In \emph{Proc. of NeurIPS}, pages 3311--3319.

\bibitem[{Berg-Kirkpatrick et~al.(2010)Berg-Kirkpatrick, Bouchard-C{\^o}t{\'e},
  DeNero, and Klein}]{berg2010painless}
Taylor Berg-Kirkpatrick, Alexandre Bouchard-C{\^o}t{\'e}, John DeNero, and Dan
  Klein. 2010.
\newblock \href {https://aclanthology.org/N10-1083} {Painless unsupervised
  learning with features}.
\newblock In \emph{Proc. of NAACL-HLT}, pages 582--590, Los Angeles,
  California.

\bibitem[{Bohnet et~al.(2018)Bohnet, McDonald, Sim{\~o}es, Andor, Pitler, and
  Maynez}]{bohnet-etal-2018-morphosyntactic}
Bernd Bohnet, Ryan McDonald, Gon{\c{c}}alo Sim{\~o}es, Daniel Andor, Emily
  Pitler, and Joshua Maynez. 2018.
\newblock \href {https://doi.org/10.18653/v1/P18-1246} {Morphosyntactic tagging
  with a meta-{B}i{LSTM} model over context sensitive token encodings}.
\newblock In \emph{Proc. of ACL}, pages 2642--2652, Melbourne, Australia.

\bibitem[{Cai et~al.(2017)Cai, Jiang, and Tu}]{cai-etal-2017-crf}
Jiong Cai, Yong Jiang, and Kewei Tu. 2017.
\newblock \href {https://doi.org/10.18653/v1/D17-1171} {{CRF} autoencoder for
  unsupervised dependency parsing}.
\newblock In \emph{Proc. of EMNLP}, pages 1638--1643, Copenhagen, Denmark.

\bibitem[{Cao et~al.(2020)Cao, Kitaev, and Klein}]{cao-etal-2020-unsupervised}
Steven Cao, Nikita Kitaev, and Dan Klein. 2020.
\newblock \href {https://doi.org/10.18653/v1/2020.emnlp-main.389} {Unsupervised
  parsing via constituency tests}.
\newblock In \emph{Proc. of EMNLP}, pages 4798--4808, Online.

\bibitem[{Che et~al.(2018)Che, Liu, Wang, Zheng, and
  Liu}]{che-etal-2018-towards}
Wanxiang Che, Yijia Liu, Yuxuan Wang, Bo~Zheng, and Ting Liu. 2018.
\newblock \href {https://doi.org/10.18653/v1/K18-2005} {Towards better {UD}
  parsing: Deep contextualized word embeddings, ensemble, and treebank
  concatenation}.
\newblock In \emph{Proc. of {C}o{NLL}}, pages 55--64, Brussels, Belgium.

\bibitem[{Christodoulopoulos et~al.(2010)Christodoulopoulos, Goldwater, and
  Steedman}]{christodoulopoulos-etal-2010-two}
Christos Christodoulopoulos, Sharon Goldwater, and Mark Steedman. 2010.
\newblock \href {https://aclanthology.org/D10-1056} {Two decades of
  unsupervised {POS} induction: How far have we come?}
\newblock In \emph{Proc. of EMNLP}, pages 575--584, Cambridge, MA.

\bibitem[{Cross and Huang(2016)}]{cross-huang-2016-span}
James Cross and Liang Huang. 2016.
\newblock \href {https://doi.org/10.18653/v1/D16-1001} {Span-based constituency
  parsing with a structure-label system and provably optimal dynamic oracles}.
\newblock In \emph{Proc. of EMNLP}, pages 1--11, Austin, Texas.

\bibitem[{Devlin et~al.(2019)Devlin, Chang, Lee, and
  Toutanova}]{devlin-etal-2019-bert}
Jacob Devlin, Ming-Wei Chang, Kenton Lee, and Kristina Toutanova. 2019.
\newblock \href {https://doi.org/10.18653/v1/N19-1423} {{BERT}: Pre-training of
  deep bidirectional transformers for language understanding}.
\newblock In \emph{Proc. of NAACL-HLT}, pages 4171--4186, Minneapolis,
  Minnesota.

\bibitem[{Dozat and Manning(2017)}]{Timothy-d17-biaffine}
Timothy Dozat and Christopher~D. Manning. 2017.
\newblock \href {https://openreview.net/forum?id=Hk95PK9le} {Deep biaffine
  attention for neural dependency parsing}.
\newblock In \emph{Proc. of ICLR}.

\bibitem[{Ethayarajh(2019)}]{ethayarajh-2019-contextual}
Kawin Ethayarajh. 2019.
\newblock \href {https://doi.org/10.18653/v1/D19-1006} {How contextual are
  contextualized word representations? comparing the geometry of {BERT},
  {ELM}o, and {GPT}-2 embeddings}.
\newblock In \emph{Proc. of EMNLP}, pages 55--65, Hong Kong, China.

\bibitem[{Gra{\c{c}}a et~al.(2009)Gra{\c{c}}a, Ganchev, Taskar, and
  Pereira}]{ganchev-etal-2009-posterior}
Jo{\~{a}}o Gra{\c{c}}a, Kuzman Ganchev, Ben Taskar, and Fernando C.~N. Pereira.
  2009.
\newblock \href
  {https://proceedings.neurips.cc/paper/2009/hash/8f1d43620bc6bb580df6e80b0dc05c48-Abstract.html}
  {Posterior vs parameter sparsity in latent variable models}.
\newblock In \emph{Proc. of NeurIPS}, pages 664--672.

\bibitem[{Gupta et~al.(2020)Gupta, Shi, Gimpel, and
  Sachan}]{gupta2020clustering}
Vikram Gupta, Haoyue Shi, Kevin Gimpel, and Mrinmaya Sachan. 2020.
\newblock \href {https://arxiv.org/abs/2010.12784} {Clustering contextualized
  representations of text for unsupervised syntax induction}.
\newblock \emph{ArXiv preprint}, abs/2010.12784.

\bibitem[{Han et~al.(2017)Han, Jiang, and Tu}]{han-etal-2017-dependency}
Wenjuan Han, Yong Jiang, and Kewei Tu. 2017.
\newblock \href {https://doi.org/10.18653/v1/D17-1176} {Dependency grammar
  induction with neural lexicalization and big training data}.
\newblock In \emph{Proc. of EMNLP}, pages 1683--1688, Copenhagen, Denmark.

\bibitem[{He et~al.(2018)He, Neubig, and Berg-Kirkpatrick}]{he2018unsupervised}
Junxian He, Graham Neubig, and Taylor Berg-Kirkpatrick. 2018.
\newblock \href {https://doi.org/10.18653/v1/D18-1160} {Unsupervised learning
  of syntactic structure with invertible neural projections}.
\newblock In \emph{Proc. of EMNLP}, pages 1292--1302, Brussels, Belgium.

\bibitem[{Huang et~al.(2015)Huang, Xu, and Yu}]{huang-etal-2015-bidirectional}
Zhiheng Huang, Wei Xu, and Kai Yu. 2015.
\newblock \href {https://arxiv.org/abs/1508.01991} {Bidirectional {LSTM-CRF}
  models for sequence tagging}.
\newblock \emph{ArXiv preprint}, abs/1508.01991.

\bibitem[{Jia et~al.(2020)Jia, Ma, Cai, and Tu}]{jia-etal-2020-semi}
Zixia Jia, Youmi Ma, Jiong Cai, and Kewei Tu. 2020.
\newblock \href {https://doi.org/10.18653/v1/2020.acl-main.607}
  {Semi-supervised semantic dependency parsing using {CRF} autoencoders}.
\newblock In \emph{Proc. of ACL}, pages 6795--6805, Online.

\bibitem[{Kim et~al.(2020)Kim, Choi, Edmiston, and Lee}]{kim-etal-2020-are}
Taeuk Kim, Jihun Choi, Daniel Edmiston, and Sang{-}goo Lee. 2020.
\newblock \href {https://openreview.net/forum?id=H1xPR3NtPB} {Are pre-trained
  language models aware of phrases? simple but strong baselines for grammar
  induction}.
\newblock In \emph{Proc. of ICLR}.

\bibitem[{Klein and Manning(2004)}]{klein-etal-2004-corpus}
Dan Klein and Christopher Manning. 2004.
\newblock \href {https://doi.org/10.3115/1218955.1219016} {Corpus-based
  induction of syntactic structure: Models of dependency and constituency}.
\newblock In \emph{Proc. of ACL}, pages 478--485, Barcelona, Spain.

\bibitem[{Li and Eisner(2019)}]{li-eisner-2019-specializing}
Xiang~Lisa Li and Jason Eisner. 2019.
\newblock \href {https://doi.org/10.18653/v1/D19-1276} {Specializing word
  embeddings (for parsing) by information bottleneck}.
\newblock In \emph{Proc. of EMNLP}, pages 2744--2754, Hong Kong, China.

\bibitem[{Liang et~al.(2006)Liang, Taskar, and
  Klein}]{liang-etal-2006-alignment}
Percy Liang, Ben Taskar, and Dan Klein. 2006.
\newblock \href {https://aclanthology.org/N06-1014} {Alignment by agreement}.
\newblock In \emph{Proc. of NAACL-HLT}, pages 104--111, New York City, USA.

\bibitem[{Lin et~al.(2015)Lin, Ammar, Dyer, and Levin}]{lin2015unsupervised}
Chu-Cheng Lin, Waleed Ammar, Chris Dyer, and Lori Levin. 2015.
\newblock \href {https://doi.org/10.3115/v1/N15-1144} {Unsupervised {POS}
  induction with word embeddings}.
\newblock In \emph{Proc. of NAACL-HLT}, pages 1311--1316, Denver, Colorado.

\bibitem[{McDonald et~al.(2013)McDonald, Nivre, Quirmbach-Brundage, Goldberg,
  Das, Ganchev, Hall, Petrov, Zhang, T{\"a}ckstr{\"o}m, Bedini,
  Bertomeu~Castell{\'o}, and Lee}]{mcdonald-etal-2013-universal}
Ryan McDonald, Joakim Nivre, Yvonne Quirmbach-Brundage, Yoav Goldberg, Dipanjan
  Das, Kuzman Ganchev, Keith Hall, Slav Petrov, Hao Zhang, Oscar
  T{\"a}ckstr{\"o}m, Claudia Bedini, N{\'u}ria Bertomeu~Castell{\'o}, and
  Jungmee Lee. 2013.
\newblock \href {https://aclanthology.org/P13-2017} {{U}niversal {D}ependency
  annotation for multilingual parsing}.
\newblock In \emph{Proc. of ACL}, pages 92--97, Sofia, Bulgaria.

\bibitem[{Merialdo(1994)}]{merialdo-1994-tagging}
Bernard Merialdo. 1994.
\newblock \href {https://aclanthology.org/J94-2001} {Tagging {E}nglish text
  with a probabilistic model}.
\newblock \emph{Computational Linguistics}, 20(2):155--171.

\bibitem[{Pereira and Schabes(1992)}]{pereira-schabes-1992-inside-outside}
Fernando Pereira and Yves Schabes. 1992.
\newblock \href {https://aclanthology.org/H92-1024} {Inside-outside
  reestimation from partially bracketed corpora}.
\newblock In \emph{Speech and Natural Language: Proceedings of a Workshop Held
  at Harriman, New York, {F}ebruary 23-26, 1992}.

\bibitem[{Peters et~al.(2018)Peters, Neumann, Iyyer, Gardner, Clark, Lee, and
  Zettlemoyer}]{peters-etal-2018-deep}
Matthew~E. Peters, Mark Neumann, Mohit Iyyer, Matt Gardner, Christopher Clark,
  Kenton Lee, and Luke Zettlemoyer. 2018.
\newblock \href {https://doi.org/10.18653/v1/N18-1202} {Deep contextualized
  word representations}.
\newblock In \emph{Proc. of NAACL-HLT}, pages 2227--2237, New Orleans,
  Louisiana.

\bibitem[{Petrov et~al.(2012)Petrov, Das, and
  McDonald}]{petrov-etal-2012-universal}
Slav Petrov, Dipanjan Das, and Ryan McDonald. 2012.
\newblock \href
  {http://www.lrec-conf.org/proceedings/lrec2012/pdf/274_Paper.pdf} {A
  universal part-of-speech tagset}.
\newblock In \emph{Proc. of {LREC}}, pages 2089--2096, Istanbul, Turkey.

\bibitem[{Rosenberg and Hirschberg(2007)}]{rosenberg-hirschberg-2007-v}
Andrew Rosenberg and Julia Hirschberg. 2007.
\newblock \href {https://aclanthology.org/D07-1043} {{V}-measure: A conditional
  entropy-based external cluster evaluation measure}.
\newblock In \emph{Proc. of EMNLP}, pages 410--420, Prague, Czech Republic.

\bibitem[{Salakhutdinov et~al.(2003)Salakhutdinov, Roweis, and
  Ghahramani}]{salakhutdinov-etal-2003-optimization}
Ruslan Salakhutdinov, Sam~T. Roweis, and Zoubin Ghahramani. 2003.
\newblock \href {http://www.aaai.org/Library/ICML/2003/icml03-088.php}
  {Optimization with {EM} and expectation-conjugate-gradient}.
\newblock In \emph{Proc. of ICML}, pages 672--679.

\bibitem[{Seginer(2007)}]{seginer-2007-fast}
Yoav Seginer. 2007.
\newblock \href {https://aclanthology.org/P07-1049} {Fast unsupervised
  incremental parsing}.
\newblock In \emph{Proc. of ACL}, pages 384--391, Prague, Czech Republic.

\bibitem[{Stratos(2019)}]{stratos2019mutual}
Karl Stratos. 2019.
\newblock \href {https://doi.org/10.18653/v1/N19-1113} {Mutual information
  maximization for simple and accurate part-of-speech induction}.
\newblock In \emph{Proc. of NAACL-HLT}, pages 1095--1104, Minneapolis,
  Minnesota.

\bibitem[{Stratos et~al.(2016)Stratos, Collins, and
  Hsu}]{stratos-etal-2016-unsupervised}
Karl Stratos, Michael Collins, and Daniel Hsu. 2016.
\newblock \href {https://doi.org/10.1162/tacl_a_00096} {Unsupervised
  part-of-speech tagging with anchor hidden {M}arkov models}.
\newblock \emph{Transactions of the Association for Computational Linguistics},
  4:245--257.

\bibitem[{Tran et~al.(2016)Tran, Bisk, Vaswani, Marcu, and
  Knight}]{tran2016unsupervised}
Ke~M. Tran, Yonatan Bisk, Ashish Vaswani, Daniel Marcu, and Kevin Knight. 2016.
\newblock \href {https://doi.org/10.18653/v1/W16-5907} {Unsupervised neural
  hidden {M}arkov models}.
\newblock In \emph{Proc. of the Workshop on Structured Prediction for {NLP}},
  pages 63--71, Austin, TX.

\bibitem[{Wang and Chang(2016)}]{wang-chang-2016-graph}
Wenhui Wang and Baobao Chang. 2016.
\newblock \href {https://doi.org/10.18653/v1/P16-1218} {Graph-based dependency
  parsing with bidirectional {LSTM}}.
\newblock In \emph{Proc. of ACL}, pages 2306--2315, Berlin, Germany.

\bibitem[{Wu et~al.(2020)Wu, Chen, Kao, and Liu}]{wu-etal-2020-perturbed}
Zhiyong Wu, Yun Chen, Ben Kao, and Qun Liu. 2020.
\newblock \href {https://doi.org/10.18653/v1/2020.acl-main.383} {Perturbed
  masking: Parameter-free probing for analyzing and interpreting {BERT}}.
\newblock In \emph{Proc. of ACL}, pages 4166--4176, Online.

\bibitem[{Yatbaz et~al.(2012)Yatbaz, Sert, and
  Yuret}]{yatbaz-etal-2012-learning}
Mehmet~Ali Yatbaz, Enis Sert, and Deniz Yuret. 2012.
\newblock \href {https://aclanthology.org/D12-1086} {Learning syntactic
  categories using paradigmatic representations of word context}.
\newblock In \emph{Proc. of EMNLP}, pages 940--951, Jeju Island, Korea.

\bibitem[{Yuret et~al.(2014)Yuret, Yatbaz, and
  Sert}]{yuret-etal-2014-unsupervised}
Deniz Yuret, Mehmet~Ali Yatbaz, and Enis Sert. 2014.
\newblock \href {https://aclanthology.org/C14-1217} {Unsupervised
  instance-based part of speech induction using probable substitutes}.
\newblock In \emph{Proc. of COLING}, pages 2303--2313, Dublin, Ireland.

\bibitem[{Zhang et~al.(2017)Zhang, Jiang, Peng, Tu, and
  Goldwasser}]{zhang-etal-2017-semi}
Xiao Zhang, Yong Jiang, Hao Peng, Kewei Tu, and Dan Goldwasser. 2017.
\newblock \href {https://doi.org/10.18653/v1/D17-1179} {Semi-supervised
  structured prediction with neural {CRF} autoencoder}.
\newblock In \emph{Proc. of EMNLP}, pages 1701--1711, Copenhagen, Denmark.

\bibitem[{Zhou et~al.(2020)Zhou, Zhang, Li, and Zhang}]{zhou-etal-2020-is}
Houquan Zhou, Yu~Zhang, Zhenghua Li, and Min Zhang. 2020.
\newblock \href {https://doi.org/10.1007/978-3-030-60450-9\_15} {Is {POS}
  tagging necessary or even helpful for neural dependency parsing?}
\newblock In \emph{Proc. of NLPCC}, pages 179--191.

\end{thebibliography}
\bibliographystyle{acl_natbib}

\appendix
\section{Details of Evaluation Metrics}
\label{sec:details_of_metric}

The core issue of the unsupervised POS tagging evaluation is that we can not directly compute the tagging accuracy since the correspondence between ground truth tags and predicted tag indexes (index-to-tag mapping) is unknown and varies from model to model.
The different evaluation metrics handle this issue in different way.

\subsection{Many-to-One Accuracy (M-1)}
M-1 is the most commonly used evaluation metric.
It addresses the problem of correspondence by assigning each predicted tag index $j \in \mathcal{P}$ to its most frequent co-occurring ground truth tag $g_i \in \mathcal{G}$:
\begin{equation}
    \operatorname{M-1}(\mathbf{A}) = \sum_j \max_{g_i} \mathbf{A}_{g_i, j},
\end{equation}
where $\mathbf{A} \in \mathbb{R}^{n\times n}$ is contingency matrix and the matrix item $\mathbf{A}_{g_i, j}$ is the number of words which are annotated as a $g_i$ and predicted as a $j$ by the model to be evaluated.
This metric, obviously, allows different predicted indexes to map to the same ground truth tag\footnote{In the {WSJ-Split} data setting, the index-to-tag mapping of metrics for {WSJ-Dev} and {WSJ-Test} are both observed from {WSJ-Dev}.}.

\subsection{One-to-One Accuracy (1-1)}
\revised{
Different from M-1 that we allows a ground truth tag $g_i$ corresponding to multiple predicted indexes, 1-1 only allows one predicted index can be assigned to a ground truth tag, and vice versa. 
Calculating 1-1 is a typical assignment problem that finding a optimal bijection function $f: \mathcal{P} \to \mathcal{G}$ that maximums the correct matching count from all possible bijection functions $\mathcal{F}$:
\begin{equation}
    \operatorname{1-1}(\mathbf{A}) = \max_{f \in \mathcal{F}}\sum_{j}\mathbf{A}_{f(j), j}.
\end{equation}
In this paper we solve this assignment problem with the Hungarian algorithm\footnote{\url{https://en.wikipedia.org/wiki/Hungarian_algorithm}}.}

\subsection{Validity-Measure (VM)} 
VM \citep{rosenberg-hirschberg-2007-v} is an entropy-based measure, which do not require the index-to-tag mapping and considers two criteria: homogeneity $h$ and completeness $c$.
The homogeneity of a predicted index indicates the purity of its co-occurring ground truth tags.
The predicted index $j$ results the highest homogeneity when it only co-occur with $g_i$, i.e., $\mathbf{A}_{g_i, j} = \sum_{g_{i'}} \mathbf{A}_{g_{i'}, j}$, and has a low homogeneity it appears with different ground truth tags randomly.
The homogeneity of a model is the simply the sum of the homogeneity of all index indicates.
The completeness is symmetrical to homogeneity, merely exchanging the position of predicted indexes and ground truth tags.
VM employs the conditional entropy to measure the value of homogeneity and completeness:
\begin{align}
    H(\mathcal{G}\mid \mathcal{P}, \mathbf{A}) &= -\sum_j \sum_i \frac{\mathbf{A}_{g_i, j}}{N} \log \frac{\mathbf{A}_{g_i, j}}{\sum_{g_{i'}} \mathbf{A}_{g_{i'}, j}}, \\
    H(\mathcal{P}\mid \mathcal{G}, \mathbf{A}) &= -\sum_i \sum_j \frac{\mathbf{A}_{g_i, j}}{N} \log \frac{\mathbf{A}_{g_i, j}}{\sum_{j'} \mathbf{A}_{g_i, j'}},
\end{align}
where $\mathbf{A} \in \mathbb{R}^{n\times n}$ is contingency matrix and the matrix item $\mathbf{A}_{g_i, j}$ is the number of words which are annotated as a $g_i$ and predicted as a $j$.

To alleviate the impact of the size of the dataset and the numbers of the POS class, the conditional entropy is normalized by the entropy of ground truth POS tag $H(\mathcal{G}, \mathbf{A})$ and $H(\mathcal{P}, \mathbf{A})$ for homogeneity and completeness, respectively:
\begin{align}
    h(\mathbf{A}) &= 1 - \frac{H(\mathcal{G}\mid \mathcal{P}, \mathbf{A})}{H(\mathcal{G}, \mathbf{A})}, \\
    c(\mathbf{A}) &= 1 - \frac{H(\mathcal{P}\mid \mathcal{G}, \mathbf{A})}{H(\mathcal{P}, \mathbf{A})},
\end{align}
where
\begin{align}
    H(\mathcal{G}, \mathbf{A}) &= -\sum_i \frac{\sum_j \mathbf{A}_{g_i, j}}{N} \log \frac{\sum_j \mathbf{A}_{g_i, j}}{N}, \\
    H(\mathcal{P}, \mathbf{A}) &= -\sum_j \frac{\sum_{g_i} \mathbf{A}_{g_i, j}}{N} \log \frac{\sum_{g_i} \mathbf{A}_{g_i, j}}{N}.
\end{align}
Completeness is symmetrical to homogeneity, merely exchanging $\mathcal{G}$ and $\mathcal{P}$ in the formulas.

In order to balance the significance between homogeneity and completeness, VM is defined as the weighted harmonic mean of homogeneity and completeness:
\begin{equation}
    \operatorname{VM}(\mathbf{A}) = \frac{(1+\beta) h(\mathbf{A})c(\mathbf{A})}{\beta h(\mathbf{A}) + c(\mathbf{A})},
\end{equation}
where $\beta$ are set to $1$ in experiments.

\section{Details of Hyper-parameters}
\label{sec:details_of_hyper_parameters}
\subsection{Model}
The number of predicted POS tags is $45$ for experiments on WSJ and $12$ for {Multilingual} experiments.
The ELMo parameters we use for experiments on WSJ are ``\texttt{Original (5.5B)}''\footnote{\url{https://allennlp.org/elmo}} from AllenNLP.
The parameters for {Multilingual} are from ``\texttt{ELMoForManyLangs}''\footnote{\url{https://github.com/HIT-SCIR/ELMoForManyLangs}} \citep{che-etal-2018-towards}.
We use ``\texttt{bert-base-cased}'' (BERT) and ``\texttt{bert-base-multilingual-cased}'' (mBERT)\footnote{\url{https://github.com/google-research/bert}} for the ablation study of PLMs on WSJ and UD respectively.
We do not fine-tune ELMo parameters.
The dropout value is uniformly set to $0.33$, and the negative slope of the activation function Leaky-ReLU is set to $1\e{-2}$.
The seeds we selected for experiments are ${0, 1, 2, 3, 4}$.

\subsection{Feature}
We set the feature cutoff threshold to $50$, which means that all features that appear in the training data less than $50$ times are replaced with a special ``\texttt{UNK}'' feature.

\subsection{Training}
We use a mini-batch update strategy with a batch size of 5000 words and optimize models with Adam.
The learning rate used in the training of the FHMM in the first step is $0.5$.
The CRF encoder is then trained on pseudo-labeled data for $5$ epochs with a learning rate of $2\e{-3}$ in the subsequent pre-training step.
In the final step, the CRF encoder has a learning rate of $1\e{-2}$, and we set the reconstruction learning rate to $2\e{-1}$.
Other hyper-parameters are identical among all three steps in training procedure,
The $\beta_1$ and $\beta_2$ are both $0.9$.
The learning rate decay is $0.75$ per $45$ epochs, the gradient clipping value is $5$, and the weight decay value $\lambda$ is $1\e{-5}$.

\section{Details of Re-run INP-GHMM Hyper-parameters}
\label{sec:details_of_inp_ghmm_hyper_parameters}
We use the word-wise character-level convolutional layer (0-th layer) of ELMo to extract word embeddings.
We use $8$ coupling layers.
To accelerate the training of INP-GHMM, we increase the batch sizes from $32$ to $512$ sentences.
We also decrease the learning rate from $1\e{-3}$ to $5\e{-4}$, as we found that high learning rates lead to performance decreases as training progresses.

\begin{table}[tb]
    \setlength{\tabcolsep}{4.7pt}
    \centering
    \begin{tabular}{lc|cc|r}
    \toprule
                               & Gender & Singular         & Plural           &  Gloss                        \\ 
    \midrule
    \multirow{2}{*}{Adj.} & M. & \lf{ross}\textbf{o}   & \lf{ross}\textbf{i}   & \multirow{2}{*}{red}          \\
                               & F.  & \lf{ross}\textbf{a}   & \lf{ross}\textbf{e}   &                               \\
    \hline
    \multirow{2}{*}{Pron.}\topsh   & M. & \lf{l}\textbf{o}      & \lf{l}\textbf{i}      & him/                          \\
                               & F.  & \lf{l}\textbf{a}      & \lf{l}\textbf{e}      & her/them                      \\ 
    \hline
    \multirow{2}{*}{Noun}\topsh      & M. & \lf{bambin}\textbf{o} & \lf{bambin}\textbf{i} & \multirow{2}{*}{boy/girl}     \\
                               & F.  & \lf{bambin}\textbf{a} & \lf{bambin}\textbf{e} &                               \\ 
    \bottomrule
    \end{tabular}
    \caption{Examples of inflections of Italian adjective, pronoun, and noun. ``M.'' means the gender Masculine and ``F.'' means Feminine.
    } 
    \label{table:inflections_example}
\end{table}

\begin{table}[tb]
    \setlength{\tabcolsep}{7.75pt}
    \centering
    \begin{tabular}{l|lll}
    \toprule
        Langs.                  & Uni-gram      & Bi-gram       & Tri-gram       \\ 
    \midrule 
        \multirow{2}{*}{it}      & \lf{mus}\bt{e}\ls{o}     & \lf{mu}\bt{se}\ls{o}     & \lf{m}\bt{use}\ls{o}     \\
                                 & \lf{mus}\bt{e}\ls{i}     & \lf{mu}\bt{se}\ls{i}     & \lf{m}\bt{use}\ls{i}     \\
    \hline
        \multirow{2}{*}{de}\topsh& \lf{mus}\bt{e}\ls{um}    & \lf{mu}\bt{se}\ls{um}    & \lf{m}\bt{use}\ls{um}    \\
                                 & \lf{mus}\bt{e}\ls{en}    & \lf{mu}\bt{se}\ls{en}    & \lf{m}\bt{use}\ls{en}    \\
    \hline
        \multirow{2}{*}{fr}\topsh& \lf{mus}\bt{\'e}\ls{e}   & \lf{mu}\bt{s\'e}\ls{e}   & \lf{m}\bt{us\'e}\ls{e}   \\
                                 & \lf{mus}\bt{\'e}\ls{e}\s & \lf{mu}\bt{s\'e}\ls{e}\s & \lf{m}\bt{us\'e}\ls{e}\s \\
        \multirow{2}{*}{es}      & \lf{mus}\bt{e}\ls{o}     & \lf{mu}\bt{se}\ls{o}     & \lf{m}\bt{use}\ls{o}     \\
                                 & \lf{mus}\bt{e}\ls{o}\s   & \lf{mu}\bt{se}\ls{o}\s   & \lf{m}\bt{use}\ls{o}\s   \\
        \multirow{2}{*}{pt-br}   & \lf{mus}\bt{e}\ls{u}     & \lf{mu}\bt{se}\ls{u}     & \lf{m}\bt{use}\ls{u}     \\
                                 & \lf{mus}\bt{e}\ls{u}\s   & \lf{mu}\bt{se}\ls{u}\s   & \lf{m}\bt{use}\ls{u}\s   \\
    \hline
        \multirow{2}{*}{en}\topsh& \lf{museu}\bt{m}         & \lf{muse}\bt{um}         & \lf{mus}\bt{eum}         \\
                                 & \lf{museu}\bt{m}\s       & \lf{muse}\bt{um}\s       & \lf{mus}\bt{eum}\s       \\
    \bottomrule
    \end{tabular}
    \caption{Language-specific suffix features for the UD datasets. The underlined \textbf{\ul{characters}} represent extracted suffix features. 
    } 
    \label{table:feature_example_ud}
\end{table}

\section{Explanation of adjustments for specific languages on UD}
Most of European languages are inflected languages.
Some words are inflected for number, gender, tense, aspect and so on.
For example in English nouns are inflected for number (singular or plural); verbs for tense.
A major way to inflect words is adding inflectional suffixes to the end of words, e.g., English nouns inflected for number with suffix ``s'' (``museum'' $\rightarrow$ ``museum\textbf{s}'').
Therefore, in some languages suffixes is more closely related to inflections than coarse-grained POS.
For instance, as shown in Table~\ref{table:inflections_example}, the last letter of Italian words is highly corresponding to gender and number, and haves little connection to coarse-grained POS.
In this work, we simply remove a certain number of ending characters for a word form before extracting suffix features, as shown in Table~\ref{table:feature_example_ud}.

\begin{table*}[tb]
    \centering
    \renewcommand{\arraystretch}{1.35}
    \setlength{\tabcolsep}{0.35pt}
    \begin{tabular}{l|*{11}c}
        \toprule
        \rowcolor[gray]{.9}
        UD-Dev                                   & de$^\ast$                         & en$^\ast$                       & es$^\ast$                       & fr$^\ast$                       & id                              & it$^\ast$                       & ja                              & ko                              & pt-br$^\ast$                    & sv                              & Mean           \\
        \midrule
        \namebox{Full CRF-AE}                    & \resultboxud{68.52}{2.1}          & \resultboxud{\textbf{68.64}}{3.0} & \resultboxud{\textbf{69.55}}{1.9} & \resultboxud{\textbf{70.30}}{0.3} & \resultboxud{\textbf{61.94}}{0.8} & \resultboxud{\textbf{69.77}}{1.6} & \resultboxud{47.91}{1.4}          & \resultboxud{36.61}{2.2}          & \resultboxud{\textbf{68.14}}{2.5} & \resultboxud{\textbf{66.99}}{2.9} & \namebox{\textbf{62.84}} \\
        \rowcolor[gray]{.95}
        \namebox{\quad w/o Features}             & \resultboxud{60.12}{1.3}          & \resultboxud{61.11}{0.8}          & \resultboxud{64.11}{4.0}          & \resultboxud{64.28}{0.6}          & \resultboxud{54.26}{2.5}          & \resultboxud{60.40}{1.5}          & \resultboxud{35.63}{3.1}          & \resultboxud{24.29}{3.0}          & \resultboxud{57.72}{3.5}          & \resultboxud{59.98}{4.2}          & \namebox{54.19} \\
        \namebox{\quad w/o UD Adjust.}           & \resultboxud{64.30}{2.0}          & \resultboxud{60.13}{2.7}          & \resultboxud{66.68}{3.4}          & \resultboxud{64.10}{4.0}          & \resultboxud{56.45}{1.7}          & \resultboxud{60.98}{3.2}          & \resultboxud{42.05}{1.9}          & \resultboxud{37.48}{2.4}          & \resultboxud{60.64}{2.6}          & \resultboxud{57.98}{1.9}          & \namebox{57.08} \\
        \rowcolor[gray]{.95}
        \namebox{\quad w/o Language Adjust.\ \ } & \resultboxud{67.26}{1.5}          & \resultboxud{67.85}{1.4}          & \resultboxud{65.83}{2.5}          & \resultboxud{66.63}{2.3}          & \namebox{$\Leftrightline$}      & \resultboxud{60.90}{6.9}          & \namebox{$\Leftrightline$}      & \namebox{$\Leftrightline$}      & \resultboxud{58.53}{1.8}          & \namebox{$\Leftrightline$}      & \namebox{60.05} \\
        \namebox{\quad w/o PLM Repr.}            & \resultboxud{65.36}{1.9}          & \resultboxud{64.28}{1.5}          & \resultboxud{64.18}{1.6}          & \resultboxud{67.76}{1.3}          & \resultboxud{56.26}{1.3}          & \resultboxud{61.64}{1.8}          & \resultboxud{42.15}{1.4}          & \resultboxud{26.29}{1.0}          & \resultboxud{64.21}{2.6}          & \resultboxud{58.27}{1.0}          & \namebox{57.04} \\
        \rowcolor[gray]{.95}
        \namebox{\quad w/o 3-stage Training}     & \resultboxud{\textbf{69.37}}{2.3} & \resultboxud{61.44}{3.8}          & \resultboxud{65.31}{1.9}          & \resultboxud{66.52}{2.8}          & \resultboxud{61.02}{2.9}          & \resultboxud{61.95}{1.4}          & \resultboxud{44.78}{2.3}          & \resultboxud{\textbf{38.96}}{1.9} & \resultboxud{66.11}{3.2}          & \resultboxud{61.72}{4.2}          & \namebox{59.72} \\
        \namebox{ELMo $\to$ mBERT}               & \resultboxud{67.54}{2.5}          & \resultboxud{66.71}{1.6}          & \resultboxud{66.74}{2.5}          & \resultboxud{68.42}{0.7}          & \resultboxud{60.55}{0.8}          & \resultboxud{65.79}{2.0}          & \resultboxud{\textbf{48.39}}{1.7} & \resultboxud{34.39}{2.1}          & \resultboxud{66.45}{2.6}          & \resultboxud{64.41}{1.3}          & \namebox{60.94} \\
        \midrule
        \midrule
        \rowcolor[gray]{.9}
        UD-Dev                                   & de                                & en                       & es                       & fr                       & id                              & it                       & ja                              & ko                              & pt-br                    & sv                              & Mean           \\
        \midrule
        \namebox{HMM (re-Impl.)}                 & \resultboxud{40.86}{1.1}          & \resultboxud{45.01}{1.8}          & \resultboxud{46.99}{2.2}          & \resultboxud{50.24}{2.1}          & \resultboxud{37.41}{1.8}          & \resultboxud{47.68}{2.3}          & \resultboxud{27.31}{1.0}          & \resultboxud{8.32}{0.8}           & \resultboxud{38.33}{4.2}          & \resultboxud{41.77}{4.3}          & \namebox{38.39} \\
        \rowcolor[gray]{.95}
        \namebox{FHMM (re-Impl.)}                & \resultboxud{58.25}{1.3}          & \resultboxud{62.75}{1.7}          & \resultboxud{60.37}{2.2}          & \resultboxud{63.68}{1.2}          & \resultboxud{49.32}{0.7}          & \resultboxud{60.51}{1.1}          & \resultboxud{34.09}{0.8}          & \resultboxud{21.48}{1.7}          & \resultboxud{60.47}{2.0}          & \resultboxud{55.41}{1.0}          & \namebox{52.63} \\
        \namebox{GHMM (re-Run)}                  & \resultboxud{72.61}{1.5}          & \resultboxud{61.34}{1.8}          & \resultboxud{68.47}{3.2}          & \resultboxud{62.46}{5.6}          & \resultboxud{55.48}{4.5}          & \resultboxud{62.90}{3.3}          & \resultboxud{38.88}{0.9}          & \resultboxud{30.00}{2.1}          & \resultboxud{60.77}{4.1}          & \resultboxud{62.42}{2.4}          & \namebox{57.53} \\
        \rowcolor[gray]{.95}
        \namebox{INP-GHMM (re-Run)}              & \resultboxud{73.68}{0.8}          & \resultboxud{61.72}{2.1}          & \resultboxud{68.76}{3.5}          & \resultboxud{62.69}{5.0}          & \resultboxud{56.91}{3.1}          & \resultboxud{64.27}{4.0}          & \resultboxud{39.09}{1.0}          & \resultboxud{30.91}{1.9}          & \resultboxud{61.84}{4.0}          & \resultboxud{65.16}{2.6}          & \namebox{58.50}        \\
        \hline
        \namebox{Ours}\topsh                     & \resultboxud{69.55}{2.2}          & \resultboxud{\textbf{67.97}}{2.7} & \resultboxud{\textbf{69.05}}{1.9} & \resultboxud{\textbf{69.05}}{0.6} & \resultboxud{\textbf{62.08}}{0.5} & \resultboxud{\textbf{70.42}}{1.6} & \resultboxud{\textbf{44.23}}{1.7} & \resultboxud{\textbf{39.50}}{2.8} & \resultboxud{\textbf{67.65}}{2.3} & \resultboxud{\textbf{68.36}}{2.5} & \namebox{\textbf{62.79}} \\
        \rowcolor[gray]{.95}
        \namebox{Ours (GHMM Init.)}              & \resultboxud{\textbf{76.17}}{2.7} & \namebox{--}                      & \namebox{--}                      & \namebox{--}                      & \namebox{--}                      & \namebox{--}                      & \namebox{--}                      & \namebox{--}                      & \namebox{--}                      & \namebox{--}                      & \namebox{--} \\
        \bottomrule
    \end{tabular}
    \caption{VM results on UD-Split. \textbf{Upper Part:} The contribution of different components on UD-Dev by removing one component at a time. $\ast$ means adopting the language-specific suffix features for this language. ``$\Leftrightline$'' means the result is identical to that of Full CRF-AE. \textbf{Lower Part:} Performance comparison on UD-Test.}
    \label{table:multilingual_new_settings_vm}
\end{table*}

\begin{table*}[tb]
    \centering
    \renewcommand{\arraystretch}{1.35}
    \setlength{\tabcolsep}{0.35pt}
    \begin{tabular}{l|*{11}c}
        \toprule
        \rowcolor[gray]{.9}
        UD-Dev                                   & de$^\ast$                       & en$^\ast$                       & es$^\ast$                       & fr$^\ast$                       & id                              & it$^\ast$                       & ja                              & ko                              & pt-br$^\ast$                    & sv                              & Mean           \\
        \midrule
        \namebox{Full CRF-AE}                    & \resultboxud{66.86}{1.8}          & \resultboxud{\textbf{71.36}}{5.4} & \resultboxud{58.69}{5.0}          & \resultboxud{63.32}{3.0}          & \resultboxud{60.03}{1.3}          & \resultboxud{\textbf{65.37}}{0.6} & \resultboxud{43.01}{2.4}          & \resultboxud{\textbf{35.53}}{2.0} & \resultboxud{\textbf{64.13}}{5.8} & \resultboxud{\textbf{66.81}}{6.2} & \namebox{\textbf{59.51}} \\
        \rowcolor[gray]{.95}
        \namebox{\quad w/o Features}             & \resultboxud{57.43}{4.2}          & \resultboxud{61.98}{3.5}          & \resultboxud{59.52}{4.4}          & \resultboxud{62.20}{0.7}          & \resultboxud{52.29}{5.3}          & \resultboxud{61.75}{5.9}          & \resultboxud{36.50}{3.3}          & \resultboxud{29.37}{3.9}          & \resultboxud{52.35}{6.0}          & \resultboxud{58.34}{6.2}          & \namebox{53.17} \\
        \namebox{\quad w/o UD Adjust.}           & \resultboxud{62.33}{3.3}          & \resultboxud{58.76}{4.7}          & \resultboxud{\textbf{63.07}}{5.9} & \resultboxud{58.29}{7.4}          & \resultboxud{52.90}{3.8}          & \resultboxud{59.74}{4.7}          & \resultboxud{35.45}{2.7}          & \resultboxud{38.84}{2.9}          & \resultboxud{57.38}{2.8}          & \resultboxud{57.80}{7.8}          & \namebox{54.46} \\
        \rowcolor[gray]{.95}
        \namebox{\quad w/o Language Adjust.\ \ } & \resultboxud{63.24}{2.6}          & \resultboxud{69.83}{4.5}          & \resultboxud{55.61}{6.8}          & \resultboxud{61.90}{3.6}          & \namebox{$\Leftrightline$}        & \resultboxud{54.91}{8.1}          & \namebox{$\Leftrightline$}        & \namebox{$\Leftrightline$}        & \resultboxud{50.37}{6.6}          & \namebox{$\Leftrightline$}        & \namebox{56.12} \\
        \namebox{\quad w/o PLM Repr.}            & \resultboxud{64.68}{3.2}          & \resultboxud{69.08}{3.0}          & \resultboxud{55.64}{5.5}          & \resultboxud{\textbf{64.21}}{1.5} & \resultboxud{53.96}{3.0}          & \resultboxud{63.07}{3.7}          & \resultboxud{40.92}{1.3}          & \resultboxud{28.62}{1.5}          & \resultboxud{61.52}{5.2}          & \resultboxud{61.32}{4.1}          & \namebox{56.30} \\
        \rowcolor[gray]{.95}
        \namebox{\quad w/o 3-stage Training}     & \resultboxud{\textbf{68.49}}{3.4} & \resultboxud{62.57}{5.9}          & \resultboxud{56.15}{5.8}          & \resultboxud{61.02}{6.9}          & \resultboxud{\textbf{61.30}}{6.4} & \resultboxud{56.33}{3.5}          & \resultboxud{39.72}{2.5}          & \resultboxud{38.30}{3.3}          & \resultboxud{60.97}{5.5}          & \resultboxud{59.09}{4.8}          & \namebox{56.39} \\
        \namebox{ELMo $\to$ mBERT}               & \resultboxud{64.31}{3.9}          & \resultboxud{70.74}{2.7}          & \resultboxud{55.95}{4.9}          & \resultboxud{62.59}{1.1}          & \resultboxud{56.70}{2.4}          & \resultboxud{60.43}{1.4}          & \resultboxud{\textbf{43.15}}{2.0} & \resultboxud{33.71}{4.4}          & \resultboxud{59.55}{5.0}          & \resultboxud{66.29}{2.5}          & \namebox{57.34} \\
        \midrule
        \midrule
        \rowcolor[gray]{.9}
        UD-Dev                                   & de                       & en                       & es                       & fr                       & id                              & it                       & ja                              & ko                              & pt-br                    & sv                              & Mean           \\
        \midrule
        \namebox{HMM (re-Impl.)}                 & \resultboxud{42.06}{2.3}          & \resultboxud{50.10}{1.9}          & \resultboxud{50.48}{3.5}          & \resultboxud{52.08}{2.4}          & \resultboxud{36.70}{4.7}          & \resultboxud{46.08}{2.1}          & \resultboxud{28.98}{4.6}          & \resultboxud{23.43}{1.3}          & \resultboxud{38.35}{3.5}          & \resultboxud{42.95}{4.5}          & \namebox{41.12} \\
        \rowcolor[gray]{.95}
        \namebox{FHMM (re-Impl.)}                & \resultboxud{61.83}{3.3}          & \resultboxud{68.62}{3.4}          & \resultboxud{57.22}{6.4}          & \resultboxud{62.36}{1.8}          & \resultboxud{52.92}{1.8}          & \resultboxud{62.24}{2.0}          & \resultboxud{\textbf{41.16}}{1.9} & \resultboxud{28.49}{2.3}          & \resultboxud{62.99}{4.2}          & \resultboxud{61.17}{2.6}          & \namebox{55.90} \\
        \namebox{GHMM (re-Run)}                  & \resultboxud{67.81}{2.7}          & \resultboxud{53.41}{4.0}          & \resultboxud{\textbf{61.81}}{4.3} & \resultboxud{55.41}{6.2}          & \resultboxud{45.46}{5.7}          & \resultboxud{55.16}{4.8}          & \resultboxud{37.95}{1.7}          & \resultboxud{29.55}{4.8}          & \resultboxud{53.52}{6.5}          & \resultboxud{56.14}{6.0}          & \namebox{51.62} \\
        \rowcolor[gray]{.95}
        \namebox{INP-GHMM (re-Run)}              & \resultboxud{68.33}{1.6}          & \resultboxud{54.03}{4.4}          & \resultboxud{61.63}{5.0}          & \resultboxud{56.68}{5.0}          & \resultboxud{46.39}{4.1}          & \resultboxud{57.61}{5.3}          & \resultboxud{38.38}{2.5}          & \resultboxud{30.79}{3.8}          & \resultboxud{53.42}{5.9}          & \resultboxud{57.98}{6.5}          & \namebox{52.52} \\
        \hline
        \namebox{Ours}\topsh                     & \resultboxud{67.55}{2.5}          & \resultboxud{\textbf{70.28}}{5.2} & \resultboxud{59.82}{5.0}          & \resultboxud{\textbf{64.39}}{2.3} & \resultboxud{\textbf{60.01}}{1.3} & \resultboxud{\textbf{65.58}}{0.7} & \resultboxud{41.07}{2.8}          & \resultboxud{\textbf{34.04}}{3.1} & \resultboxud{\textbf{63.39}}{5.6} & \resultboxud{\textbf{67.63}}{6.2} & \namebox{\textbf{59.38}} \\
        \rowcolor[gray]{.95}
        \namebox{Ours (GHMM Init.)}              & \resultboxud{\textbf{71.94}}{4.0} & \namebox{--}                      & \namebox{--}                      & \namebox{--}                      & \namebox{--}                      & \namebox{--}                      & \namebox{--}                      & \namebox{--}                      & \namebox{--}                      & \namebox{--}                      & \namebox{--} \\
        \bottomrule
    \end{tabular}
    \caption{1-1 accuracy on UD-Split. \textbf{Upper Part:} The contribution of different components on UD-Dev by removing one component at a time. $\ast$ means adopting the language-specific suffix features for this language. ``$\Leftrightline$'' means the result is identical to that of Full CRF-AE. \textbf{Lower Part:} Performance comparison on UD-Test. }
    \label{table:multilingual_new_settings_oo}
\end{table*}

\begin{table*}[tb]
    \centering
    \renewcommand{\arraystretch}{1.35}
    \setlength{\tabcolsep}{0.6pt}
    \begin{tabular}{l|*{11}c}
        \toprule
        \rowcolor[gray]{.9}
        Model                           & de                                & en                       & es                       & fr                       & id                              & it                       & ja                              & ko                              & pt-br                    & sv                              & Mean           \\
        \midrule
        \namebox{Brown (C'10)}          & \namebox{60.0}                    & \namebox{62.9}                    & \namebox{67.4}                    & \namebox{66.4}                    & \namebox{59.3}                    & \namebox{66.1}                    & \namebox{60.3}                    & \namebox{47.5}                    & \namebox{67.4}                    & \namebox{61.9}                    & \namebox{61.9}           \\
        \rowcolor[gray]{.95}
        \namebox{FHMM (B'10)}           & \resultboxud{67.5}{1.8}           & \resultboxud{62.4}{3.5}           & \resultboxud{67.1}{3.1}           & \resultboxud{62.1}{4.5}           & \resultboxud{61.3}{3.9}           & \resultboxud{52.9}{2.9}           & \resultboxud{78.2}{2.9}           & \resultboxud{60.5}{3.6}           & \resultboxud{63.2}{2.2}           & \resultboxud{56.7}{2.5}           & \namebox{63.2}           \\
        \namebox{AHMM (S'16)}           & \namebox{63.4}                    & \namebox{71.4}                    & \namebox{74.3}                    & \namebox{71.9}                    & \namebox{67.3}                    & \namebox{60.2}                    & \namebox{69.4}                    & \namebox{61.8}                    & \namebox{65.8}                    & \namebox{61.0}                    & \namebox{66.7}           \\
        \rowcolor[gray]{.95}
        \namebox{MIM (S'19)}            & \resultboxud{75.4}{1.5}           & \resultboxud{73.1}{1.7}           & \resultboxud{73.1}{1.0}           & \resultboxud{70.4}{2.9}           & \resultboxud{73.6}{1.5}           & \resultboxud{67.4}{3.3}           & \resultboxud{77.9}{0.4}           & \resultboxud{65.6}{1.2}           & \resultboxud{70.7}{2.3}           & \resultboxud{67.1}{1.5}           & \namebox{71.4}           \\
        \namebox{SyntDEC (G'20)}        & \resultboxud{81.5}{1.8}           & \resultboxud{76.5}{1.1}           & \resultboxud{78.9}{1.9}           & \resultboxud{70.7}{3.9}           & \resultboxud{76.8}{1.1}           & \resultboxud{71.7}{3.3}           & \resultboxud{84.7}{1.2}           & \resultboxud{69.7}{1.5}           & \resultboxud{77.7}{2.1}           & \resultboxud{68.8}{3.9}           & \namebox{75.7}           \\
        \rowcolor[gray]{.95}
        \namebox{GHMM (re-Run)}         & \resultboxud{82.16}{1.9}          & \resultboxud{75.31}{2.1}          & \resultboxud{80.27}{2.2}          & \resultboxud{76.59}{3.7}          & \resultboxud{76.52}{4.0}          & \resultboxud{72.78}{5.8}          & \resultboxud{79.81}{0.9}          & \resultboxud{67.52}{2.0}          & \resultboxud{74.99}{4.1}          & \resultboxud{73.60}{2.9}          & \namebox{75.96}          \\
        \namebox{INP-GHMM (re-Run)\ \ } & \resultboxud{83.48}{1.8}          & \resultboxud{76.02}{1.4}          & \resultboxud{81.68}{2.7}          & \resultboxud{77.40}{3.4}          & \resultboxud{77.72}{2.7}          & \resultboxud{72.55}{5.5}          & \resultboxud{79.41}{1.5}          & \resultboxud{68.07}{1.8}          & \resultboxud{75.27}{4.5}          & \resultboxud{74.48}{3.1}          & \namebox{76.61}          \\
        \hline
        \rowcolor[gray]{.95}
        \namebox{Ours}\topsh            & \resultboxud{82.41}{2.0}          & \resultboxud{\textbf{80.79}}{1.1} & \resultboxud{\textbf{82.65}}{2.0} & \resultboxud{\textbf{82.67}}{0.6} & \resultboxud{\textbf{81.09}}{1.3} & \resultboxud{\textbf{78.13}}{1.6} & \resultboxud{\textbf{85.52}}{1.1} & \resultboxud{\textbf{74.87}}{2.7} & \resultboxud{\textbf{79.67}}{2.4} & \resultboxud{\textbf{78.44}}{3.5} & \namebox{\textbf{80.67}} \\
        \namebox{Ours (GHMM Init.)}     & \resultboxud{\textbf{86.93}}{1.2} & \namebox{--}                      & \namebox{--}                      & \namebox{--}                      & \namebox{--}                      & \namebox{--}                      & \namebox{--}                      & \namebox{--}                      & \namebox{--}                      & \namebox{--}                      & \namebox{--} \\
        \bottomrule
    \end{tabular}
    \caption{
    M-1 accuracy on UD-All.
    C'10 is for \citet{christodoulopoulos-etal-2010-two}, 
    B'10 for \citet{berg2010painless}, 
    S'16 for \citet{stratos-etal-2016-unsupervised}, 
    S'19 for \citet{stratos2019mutual}, 
    and G'20 for \citet{gupta2020clustering}.}
    \label{table:multilingual_result}
\end{table*}

\end{CJK*}
\end{document}